\def\year{2019}\relax
\documentclass[letterpaper]{article} 
\usepackage{aaai19}  
\usepackage{times}  
\usepackage{helvet} 
\usepackage{courier}  
\usepackage[hyphens]{url}  
\usepackage{graphicx} 
\usepackage{graphicx}  
\usepackage{amsmath,amsfonts,bm}

\def\eqref#1{equation~\ref{#1}}

\def\1{\bm{1}}

\def\vc{{\bm{c}}}

\def\vh{{\bm{h}}}

\def\vx{{\bm{x}}}

\def\vz{{\bm{z}}}

\DeclareMathAlphabet{\mathsfit}{\encodingdefault}{\sfdefault}{m}{sl}
\SetMathAlphabet{\mathsfit}{bold}{\encodingdefault}{\sfdefault}{bx}{n}

\newcommand{\KL}{D_{\mathrm{KL}}}

\newcommand{\Eb}{\mathbb{E}}
\usepackage{booktabs}
\usepackage[draft]{hyperref}
\usepackage{url}
\usepackage{graphicx}
\usepackage{amssymb}
\usepackage{subfigure}
\usepackage{amsthm}
\usepackage{comment}
\usepackage{color}

\def\setstretch#1{\renewcommand{\baselinestretch}{#1}}
\title{Meta-Amortized Variational Inference and Learning}
\author{Mike Wu\thanks{Denotes equal contribution.}, Kristy Choi$^{\ast}$, Noah Goodman\thanks{Also affiliated with the Psychology Department}, Stefano Ermon\\
Computer Science Department \\
Stanford University\\
\texttt{\{wumike, kechoi, ngoodman, ermon\}@cs.stanford.edu}}

\begin{document}

\maketitle

\begin{abstract}
Despite the recent success in probabilistic modeling and their applications, generative models trained using traditional inference techniques struggle to adapt to new distributions, even when the target distribution may be closely related to the ones seen during training. In this work, we present a doubly-amortized variational inference procedure as a way to address this challenge. By sharing computation across not only a set of query inputs, but also a set of different, related probabilistic models, we learn transferable latent representations that generalize across several related distributions. In particular, given a set of distributions over images, we find the learned representations to transfer to different data transformations. We empirically demonstrate the effectiveness of our method by introducing the MetaVAE, and show that it significantly outperforms baselines on downstream image classification tasks on MNIST (10-50\%) and NORB (10-35\%).
\end{abstract}
\section{Introduction}

A wide variety of problems in machine learning (ML) can be framed as probabilistic inference in generative models. In particular, latent variable models learn representations of data that capture salient characteristics of its underlying distribution, which can then be used for downstream tasks such as classification \cite{klingler2017efficient}. While traditional inference techniques can be slow or even computationally intractable, the advent of \textit{amortized (variational) inference} allowed such methods to scale to large datasets, bringing about significant progress in generative modeling applications such as image and audio synthesis \cite{brock2018large,oord2016wavenet}, molecule generation \cite{segler2017generating}, and more.

However, as the problem domains we face become increasingly more complex and multimodal, a technical challenge arises: generative models trained using traditional inference techniques struggle to adapt to new data distributions, even when these new distributions may be \textit{closely related} to distributions seen during training. For example, variational autoencoders (VAEs) trained on the original image distributions have difficulty generalizing to small visual transformations such as
changing the position or quantity of objects in the scene. 
However, we would expect the true generative model, such as those of humans \cite{yildirim2014perception}, to be invariant to these slight modifications. Therefore, the question we aim to address is: 
how do we design an amortized inference algorithm that 
generalizes across 
related distributions to learn \textit{transferable} representations? Such features would capture the salient characteristics necessary to allow for better generalization to related, but unseen distributions at test time.

To address this question, we propose a \textit{doubly-amortized} inference procedure that amortizes computation across not only a set of query inputs, but also a \textit{set} of different, related target probabilistic models.
More precisely, we derive a new objective called the MetaELBO which serves as a variational lower bound across multiple distributions, while also incorporating a prior regularization term encouraging each generative model to match its respective data marginal.
We note that this inference model is not intended to be universal, but rather tailored to a specific family where each  probabilistic model is similar in structure. 
Inspired by meta-learning, we denote this "doubly-amortized" inference problem as \textit{meta-inference} and let a \textit{meta-distribution} refer to the probability distribution over the family of probabilistic models.


As an instantiation of our method, we introduce the MetaVAE, a VAE trained with the MetaELBO. 
Empirically, we first show three demonstrations to build intuition for meta-inference: 1) clustering, 2) compiled inference, and 3) learning sufficient statistics on exponential families. Then, we study image transformations (e.g. rotations, shearing) on MNIST digits where the MetaVAE learns representations that transfer to unseen transformations, outperforming baselines by 10-50\%. Finally, we showcase similar improvements of 10-35\% on real-world images (NORB).
While the representations learned from other generative models quickly decay in quality under more severe transformations, those of the MetaVAE preserve relevant information about the image while abstracting away unnecessary differences induced by visual manipulation. 

\section{Preliminaries}

\subsection{Exact and Approximate Inference}
Let $p(\vx, \vz)$ be a joint distribution
over a set of latent variables $\vz \in \mathcal{Z}$ and observed variables $\vx \in \mathcal{X}$. An \textit{inference query} involves computing posterior beliefs after incorporating evidence into the prior: $p(\vz|\vx) = p(\vx, \vz)/p(\vx)$. This quantity is often intractable to compute as the marginal likelihood $p(\vx) = \int_{\vz} p(\vx, \vz) d\vz$ requires integrating or summing over a potentially exponential number of configurations for $\vz$. Thus, we are forced to seek approximations.

Approximate inference techniques such as Markov Chain Monte Carlo (MCMC) sampling \cite{hastings1970monte,gelfand1990sampling} and variational inference (VI) 
\cite{jordan1999introduction,wainwright2008graphical,blei2017variational} are widely used to approximate the posterior $p(\vz|\vx)$. In VI, we introduce a family of tractable distributions $\mathcal{Q}$ parameterized by $\psi$ over the latent variables and find the member (called the approximate posterior), $q_{\psi^*} \in \mathcal{Q}$ that minimizes the Kullback-Leibler (KL) divergence between itself and the exact posterior:

\begin{equation}
q_{\psi^*}(\vz) = \arg \min_{q_{\psi}} \KL(q_{\psi}(\vz)||p (\vz|\vx))
\end{equation}

This $q_{\psi^*}(\vz)$ can serve as a proxy for the true posterior distribution. We note that the solution depends on the specific value of the observed (evidence) variables $\vx$ we are conditioning on. For notational clarity, we rewrite the variational parameters as $\psi_{\vx}$ to make explicit their dependence on $\vx$. 

One commonly needs to solve multiple inference queries of the same kind, conditioning on different values of the observed variables $\vx$ (evidence). 
Let $p_{\mathcal{D}}(\vx)$ be an empirical distribution over the observed variables $\vx \in \mathcal{X}$. Note $p_{\mathcal{D}}(\vx)$ can be different from the marginal $p(\vx)$ when the model is mis-specified. 
The average quality of the variational approximations can then be quantified by:
\begin{equation}
 \mathbb{E}_{p_{\mathcal{D}}(\vx)}\left[\max_{\psi_{\vx}} \mathbb{E}_{q_{\psi_{\vx}}(\vz)} \log \frac{p(\vx, \vz)}{q_{\psi_{\vx}}(\vz)} \right] 
\label{elbo:vi:first}
\end{equation}
where $q_{\phi_{\vx}}(\vz)$ can be viewed as an importance distribution.
In practice, $p_{\mathcal{D}}(\vx)$ is unknown but we assume access to 
a training dataset $\mathcal{D}$ of examples i.i.d. sampled from $p_{\mathcal{D}}(\vx)$ that can be used to evaluate Eq.~\ref{elbo:vi:first}.


\subsection{Amortized Variational Inference}
An alternative formulation leverages a technique known as \textit{amortization} \cite{gershman2014amortized}, which reduces the computational cost of Eq.~\ref{elbo:vi:first} by casting the per-sample optimization process as a supervised \textit{regression} task. Rather than solving for an optimal $q_{\psi^*_{\vx}}(\vz)$ for every $\vx$, we learn a single deterministic mapping $f_\phi:\mathcal{X} \rightarrow \mathcal{Q}$ to \textit{predict} 
$\psi^*_{\vx}$, or equivalently $q_{\psi^*_{\vx}}(\vz) \in \mathcal{Q}$, 
as a function of $\vx$. Often, we choose to represent $f_\phi$ as a conditional distribution, denoted by $q_\phi(\vz|\vx)$ = $f_\phi(\vx)(\vz)$ when scoring a value $\vz$.

This procedure introduces an \textit{amortization gap}, in which the less flexible parameterization of the inference model replaces the objective in Eq.~\ref{elbo:vi:first} with the following lower bound:
\begin{align}
\max_{\phi}\mathbb{E}_{p_{\mathcal{D}}(\vx)} \left[ \mathbb{E}_{q_\phi(\vz|\vx)} \log \frac{p(\vx,\vz)}{q_\phi(\vz|\vx)} \right]
\label{eq:plainamortization}
\end{align}
This gap refers to the suboptimality caused by amortizing the variational parameters over the entire training set, as opposed to optimizing for each example individually (pulling the $\max$ out of the expectation in Eq.~\ref{elbo:vi:first}).
This tradeoff in expressiveness, however, enables significant speedups.

\subsection{Learning Latent Variable Models}
So far, we have assumed that the true generative model $p(\vx,\vz)$ is given. However, we often only possess a family of possible models, $p_\theta(\vx,\vz)$ parameterized by $\theta$ and the data set of observations, $\mathcal{D}$. The challenge then, is to choose $\theta$ whose model best explains the evidence.
To do so, we maximize the log marginal likelihood of the data:
\begin{equation}
\mathbb{E}_{p_{\mathcal{D}}(\vx)} \left[\log p_\theta(\vx) \right] = \mathbb{E}_{p_{\mathcal{D}}(\vx)} \left[\log \int_{\vz} 
p_\theta(\vx,\vz)d\vz \right]
\label{eqn:marginal}
\end{equation}
As mentioned, Eq.~\ref{eqn:marginal} is intractable to evaluate. Instead, we derive the Evidence Lower Bound (ELBO) to Eq.~\ref{eqn:marginal} using $q_\phi(\vz|\vx)$ as a tractable amortized inference model:

\begin{align}
    \mathbb{E}_{p_{\mathcal{D}}}[\log p_\theta(\vx)] &\geq \mathbb{E}_{p_{\mathcal{D}}(\vx)} \left[\mathbb{E}_{q_\phi(\vz|\vx)} \left[\log \frac{ p_\theta(\vx,\vz)}{q_\phi(\vz|\vx)} \right] \right]
    \label{eqn:elbo}
\end{align}
With Eq.~\ref{eqn:elbo} as an objective, we jointly optimize the parameters of the inference and generative models: $\phi$ and $\theta$. 

We may derive an alternative formulation of Eq.~\ref{eqn:elbo}:
\begin{align}
     \mathcal{L}(\phi, \theta) &= -\KL(q_\phi(\vx, \vz) \Vert p_\theta(\vx, \vz))  \\
                                &=  -\KL(p_{\mathcal{D}}(\vx) \Vert p_\theta(\vx)) \nonumber \\ 
     & \qquad-\Eb_{p_{\mathcal{D}}}[\KL(q_\phi(\vz|\vx) \Vert p_\theta(\vz|\vx)) ] \label{eqn:elbo_alt}
\end{align}
where $q_\phi(\vx, \vz) = f_\phi(\vx)(\vz)p_{\mathcal{D}}(\vx)$. Eq.~\ref{eqn:elbo_alt} is comprised of a maximum likelihood term with a regularization penalty that encourages the generative model to have posteriors that can be easily approximated by the inference model. We will revisit this intuition once we introduce meta-amortization.

Often, $p_\theta(\vx|\vz)$ and $q_\phi(\vz|\vx)$ are parameterized by deep neural networks, which is known as a variational autoencoder, or VAE \cite{kingma2013auto}.
The latent variables $\vz$ are learned ``features" inferred by $q_\phi(\vz|\vx)$ that can be used in downstream tasks, such as clustering or classification. The VAE is popular in many real-world domains: in medical diagnosis, for example, one can infer the identity of a disease ($\vz$) from observed symptoms ($\vx$). Given a set of symptoms from a population of patients, we can fit a VAE tailored to a disease, e.g. thoracic disease \cite{mao2018deep}.

\section{Meta-Amortized Variational Inference}
\label{sec:meta}


But in practice, physicians often work with several patient populations that vary across a wide range of socioeconomic factors. For a new population, clinicians draw on prior experience from patients with similar symptoms, lowering their chances of misdiagnosis. 
We can similarly construct a generative model that captures this intuition. 
Instead of training a VAE on a new population, which would be equivalent to the physician re-learning how to diagnose an illness, we aim to share statistical strength between different patient groups to infer latent features that transfer to similar, but previously unseen populations.
We formalize this idea into a new algorithm that we call \textit{meta-amortized inference}.

Recall a (singly)-amortized inference model for $p_\theta(\vx,\vz)$
\begin{equation}
\max_{\phi}\mathbb{E}_{p_{\mathcal{D}}(\vx)} \left[ \mathbb{E}_{f_\phi(\vx)} \log \frac{p_\theta(\vx,\vz)}{f_\phi(\vx)(\vz)} \right]
\end{equation}
which approximates $p_\theta(\vz|\vx)$ for various choices of the observed variables, $\vx \sim p_{\mathcal{D}}(\vx)$. Unlike Eq.~\ref{eq:plainamortization}, we have written $q_\phi(\vz|\vx)$ in its alternate form, $f_\phi(\vx)(\vz)$. 



We are now interested in not one but a set of models,
$\mathcal{J}_\mathcal{I} = \{p_{\theta_i}(\vx,\vz), i \in \mathcal{I}\}$ where $\mathcal{I}$ is a finite set of indices.  
Crucially, (like the example above) we make a few simplifying assumptions. First, we assume that the random variables in each model have the same domains (e.g. $\mathcal{X},\mathcal{Z}$), but the relationships between the random variables may be different.
Second, we assume that for each model, we care about the same inference query $p_{\theta_i}(\vz|\vx)$. Finally, we assume to have some knowledge of typical values of the observed variables for each model in $\mathcal{J}_\mathcal{I}$: formally, we desire a set $\mathcal{M}_\mathcal{I} = \{ p_{\mathcal{D}_i}(\vx), i \in \mathcal{I} \} \subseteq \mathcal{M}$ of marginal distributions over the observed variables. 
Here, $\mathcal{M}$ denotes the set of all possible marginal distributions over $\mathcal{X}$. Let $p_{\mathcal{M}}: \mathcal{M}_\mathcal{I} \rightarrow [0,1]$ denote a distribution over $\mathcal{M}_\mathcal{I}$. For example, $p_{\mathcal{M}}$ may be uniform over a finite number of marginals. As $p_{\mathcal{M}}$ is a distribution over distributions, we refer to it as a \textit{meta-distribution}.

The naive approach to amortize over a set of models is:
\begin{equation}
\mathbb{E}_{p_{\mathcal{D}_i} \sim p_{\mathcal{M}}} \left[
\max_{\phi}\mathbb{E}_{p_{\mathcal{D}_i}(\vx)} \left[ \mathbb{E}_{f_\phi(\vx)} \log \frac{p_{\theta_i}(\vx,\vz)}{f_\phi(\vx)(\vz)} \right] \right]
\end{equation}
where we separately fit an amortized inference model for each $p_{\theta_i}(\vx,\vz)$. However, this approach is prohibitively expensive as the size of $\mathcal{M}_{\mathcal{I}}$ increases, and training across models is decoupled.
We instead propose to doubly-amortize the inference procedure as follows (we move the $\max$ out once more):
\begin{equation}
\max_{\phi} \mathbb{E}_{p_{\mathcal{D}_i} \sim p_{\mathcal{M}}} \left[
\mathbb{E}_{p_{\mathcal{D}_i}(\vx)} \left[ \mathbb{E}_{g_\phi(p_{\mathcal{D}_i}, \vx)} \log \frac{p_{\theta_i}(\vx,\vz)}{g_\phi(p_{\mathcal{D}_i},\vx)(\vz)} \right] \right] 
\label{eqn:meta1_obj}
\end{equation}
where the original regressor $f_\phi(\vx)$ is replaced by a doubly-amortized regressor $g_\phi(p_{\mathcal{D}_i},\vx)$ that takes \textit{both} the marginal distribution $p_{\mathcal{D}_i}(\vx)$ and an observation $\vx$ to return a posterior distribution. Formally, we call such a mapping, $g_\phi: \mathcal{M} \times \mathcal{X} \rightarrow \mathcal{Q}$, a \textit{meta-inference model}. This doubly-amortized inference procedure must be robust across varying marginals and evidence, generalizing over $\mathcal{M}$: a large set of sufficiently similar, previously \textit{unseen} models. 

We note that the choice of $p_{\mathcal{D}_i}(\vx)$ as input to $g_\phi$ is critical in practice. As in Eq.~\ref{eqn:elbo_alt}, a successful learning algorithm will learn generative models such as $p_{\theta_i}(\vx)$ or $p_{\theta_i}(\vx, \vz)$ that match $p_{\mathcal{D}_i}(\vx)$.
But similarly to the recent progress in wake-sleep \cite{hinton1995wake,bornschein2014reweighted,le2018revisiting}, we found that using observations from the true marginal $p_{\mathcal{D}_i}(\vx)$ led to significantly more stable training.
One may also consider alternate combinations of inputs for $p_{\mathcal{D}_i}(\vx)$, which we leave as future work.

\paragraph{Meta-Amortized Variational Bayes and Learning}
In certain settings, we are given a set of generative models $\{p_{\theta_i^*}(\vx, \vz), i \in \mathcal{I} \}$, where each model $p_{\theta_i^*}(\vx, \vz)$ with known parameters  captures a marginal distribution, $p_i(\vx) \in \mathcal{M}_{\mathcal{I}}$. 
We can then immediately optimize Eq.~\ref{eqn:meta1_obj} to obtain the optimal meta-inference model. 

But in many cases the generative models are not known ahead of time, and therefore we must jointly learn $\{\theta_i,  i \in \mathcal{I}\}$ along with the parameters of the meta-inference model, $\phi$. To do so, we consider the objective, 
\begin{equation}
\max_{\phi} \mathbb{E}_{p_{\mathcal{D}_i} \sim p_\mathcal{M}} \left[ \max_{\theta_i}  
\mathcal{L}_{\phi, \theta_i}(p_{\mathcal{D}_i}) 
\right]
\label{eqn:meta2_obj}
\end{equation}
where the inner loss function is defined as:
\begin{equation*}
    \mathcal{L}_{\phi, \theta_i}(p_{\mathcal{D}_i}) = -\KL(p_{\mathcal{D}_i}(\vx) g_\phi(p_{\mathcal{D}_i}, \vx) || p(\vz)p_{\theta_i}(\vx|\vz))
\end{equation*}
and $p_{\mathcal{D}_i}(\vx) g_\phi(p_{\mathcal{D}_i}, \vx)$ denotes the distribution defined implicitly by first sampling $\vx \sim p_i(\vx)$, then sampling $\vz \sim g_\phi(p_{\mathcal{D}_i}, \vx)$. We refer to this lower bound as the MetaELBO, and a VAE trained with this objective as the MetaVAE. 

Lastly, as we did in Eq.~\ref{eqn:elbo_alt}, we can rewrite the MetaELBO to a more interpretable form. Similar to $f_\phi(\vx)$, our regressor $g_\phi(p_{\mathcal{D}_i}, \vx)$ can be represented as a conditional distribution, denoted $q_\phi(\vz|p_{\mathcal{D}_i}, \vx) = g_\phi(p_{\mathcal{D}_i}, \vx)(\vz)$.
Then,
\begin{align*}
    \mathcal{L}_{\phi, \theta}(p_{\mathcal{D}_i}) &= -\KL(p_{\mathcal{D}_i}(\vx) q_\phi(\vz| p_{\mathcal{D}_i}, \vx) || p(\vz)p_{\theta_i}(\vx|\vz)) \\
    &= -\KL(p_{\mathcal{D}_i}(\vx)||p_{\theta_i}(\vx)) \\
    & \qquad -\mathbb{E}_{\vx \sim p_{\mathcal{D}_i}(\vx)}[\KL(q_\phi(\vz|p_{\mathcal{D}_i}, \vx)||p_{\theta_i}(\vz|\vx))].
\end{align*}
This form has a penalty term for each distribution $p_{\mathcal{D}_i}(\vx)$, encouraging the meta-amortized inference model to perform well across $p_{\mathcal{D}_i}(\vx)$ sampled from the meta-distribution $p_{\mathcal{M}}$. We note that if $\mathcal{M} = \{p_{\mathcal{D}}\}$, then $g_\phi(p_{\mathcal{D}_i}, \vx) = f_\phi(\vx)$, and the MetaELBO is equivalent to ELBO. 

Interestingly, we find that the MetaVAE's learned representations transfer well to unseen downstream tasks at test time. We provide some intuition as to why this is the case. Samples from the corresponding marginal $p_{\mathcal{D}_i}$ help to lower the variance in the meta-inference network's inferred $\vz$'s for each query point $\vx$, regularizing the model's behavior to yield more robust representations.

\subsection{Representing the Meta-Distribution}
In Eq.~\ref{eqn:meta2_obj}, it is not clear how to represent a distribution  $p_{\mathcal{D}_i}(\vx)$ as input if we parameterize $g_\phi(p_{\mathcal{D}_i}, \vx)$ as a neural network. One of the main insights from this work is to represent the marginal distribution as a finite set of samples, 
\begin{equation}
    \mathcal{D}_i = \{\vx_j \sim p_{\mathcal{D}_i}(\vx) | j=1,...,N\}
\end{equation}
or a \textit{data set}. 
We can then use $\mathcal{D}_i$ to define an empirical analogue to $g_\phi(p_i, \vx)$, denoted as $\hat{g}_\phi:\mathcal{X}^N \times \mathcal{X} \rightarrow \mathcal{Q}$, which maps a data set with $N$ samples and an observation to a posterior. Then, there is an equivalent analogue of Eq.~\ref{eqn:meta2_obj} where a marginal, $p_{\mathcal{D}_i}(\vx)$ is replaced by a data set, $\mathcal{D}_i$. 


\section{Related work}
\paragraph{Rapid Adaptation through Meta-Learning.} Among the rich body of work on meta-learning \cite{vinyals2016matching,snell2017prototypical,gordon2018decision}, a common goal is to train models such that they will rapidly adapt to new, unseen classification tasks. Although the Neural Process (NP) \cite{garnelo2018neural,kim2019attentive} is similar to our work in that it derives predictions for new targets by conditioning the encoder network on a relevant \textit{context set}, it models uncertainty over a distribution of \textit{functions}.
Another line of research formulates proper initialization as the workhorse of successful meta-learning \cite{finn2017model,grant2018recasting}. In many ways, our meta-amortized inference procedure can be thought of as learning a good initialization for an inference model on a new target distribution. However, these approaches are not directly comparable to ours because of their supervised nature. 

\paragraph{Few-shot Generative Modeling.} This branch of research aims to train generative models such that they will generalize to unseen distributions at test time given only a few examples. The focus has been on few-shot density estimation, with approaches ranging from the use of
conditioning \cite{bartunov2016fast} to nested optimization \cite{reed2017few}. Meta-inference however is not few-shot, and instead aims to learn transferable \textit{representations} for downstream tasks rather than density estimation alone.

The most relevant prior works include the Neural Statistician \cite{edwards2016towards} (NS) and the Variational Homoencoder \cite{hewitt2018variational} (VHE), two very similar models that study inference over sets of observations. The VHE optimizes the following objective, 
\begin{equation}
    \mathbb{E}_{\vx,\mathcal{D} \sim p_{\mathcal{D}}}[\mathbb{E}_{q_\phi(\vc|\mathcal{D})}[\log p_\theta(\vx|\vc)] - \frac{1}{N}\KL(q_\phi(\vc|\mathcal{D})||p(\vc))]
    \label{eqn:homoencoder}
\end{equation}
where $\mathcal{D} = \{\vx_1, ..., \vx_N\}$ is a set of $N$ samples and $\vc$ is a global latent variable. 
We note that if we view $\mathcal{D}$ as an approximation for a marginal distribution, then NS and VHE also serve as baselines that can perform doubly-amortized inference. Like our proposed inference model $\hat{g}_\phi(\mathcal{D}, \vx)$, the distribution $q(\vc|\mathcal{D})$ in Eq.~\ref{eqn:homoencoder} ingests a data set.
However, both the VHE and NS utilize a global variable $\vc$ (isotropic Gaussian). 
We believe this constraint is \textit{overly restrictive} in settings which require transferring to a diverse set of distributions, hurting generalization performance. Instead, the MetaVAE does not impose a distributional assumption on the different generative models, and we find that this non-parametric approach yields consistently better performance. 

\section{Demo: Clustering Mixtures of Gaussians}
First, we present a simple clustering example to build intuition for meta-inference. Consider a standard VAE trained to capture a single mixture of two Gaussian (MoG) distributions $p_{\mathcal{D}}(\vx)$. Each component has isotropic covariance of $0.1$ and mean drawn from the uniform distribution, $U(-5, 5)$. The two components are mixed evenly and assigned a label of 0 or 1.
Then, inference $q_\phi(\vz|\vx)$ with  $\vz \in \{0, 1\}$ as a 1-D binary latent variable amounts to predicting which component $\vx$ belongs to, of which the true cluster label is recoverable up to a permutation.

Now we introduce meta-inference for this task. 
Given that an inference model $q_\phi(\vz|\vx)$ of a VAE can learn to cluster data from a \emph{specific} MoG, a meta-inference model $g_\phi(p_{\mathcal{D}_i},\vx)$ should correspond to \textit{a general-purpose clustering algorithm} that can separate out the components of any related, but previously  unseen mixture distribution $p_{\mathcal{D}_i}$. 

Concretely, we let each distribution $p_{\mathcal{D}_i}(\vx) \sim p_\mathcal{M}$ be a MoG
and train a MetaVAE amortized over $N$ mixtures
to assess how well it can predict $\vz \in \{0,1\}$ for a given $\vx$ for an \emph{unseen test distribution}. 
We measure this clustering accuracy on 1000 unseen but related MoGs 
sampled from the same meta-train distribution.
While the VAE has a clustering error of $27.9$\% due to cases where there is extreme overlap in mixture components,
the MetaVAE has an error of 9.9\% when $N = 50$. 
Moreover, larger $N$ improved the model's performance ($21.2$\% error with $N=10$ and $15.8$\% error with $N=20$) as expected. 
We include more details and a second study on clustering MNIST digits in the Appendix.

\section{Demo: Inference for Classical Mechanics}

For a second demonstration, we consider an introductory problem in classical mechanics: objects sliding down inclined planes. Here, we are given a physics simulator that models a box that faces friction with the plane. 
Each time the simulator runs, we see a new box with a different friction coefficient. The simulator then records the time it takes for the box to descend to the bottom of the plane. 
Each simulator has a different incline plane of length $L$ and incline angle $A$, and our task is to infer the coefficient of friction ($\vz$) from the observed descent time $(\vx)$ given a new simulator. 
\begin{figure}[h!]
\centering     
\includegraphics[width=0.9\linewidth]{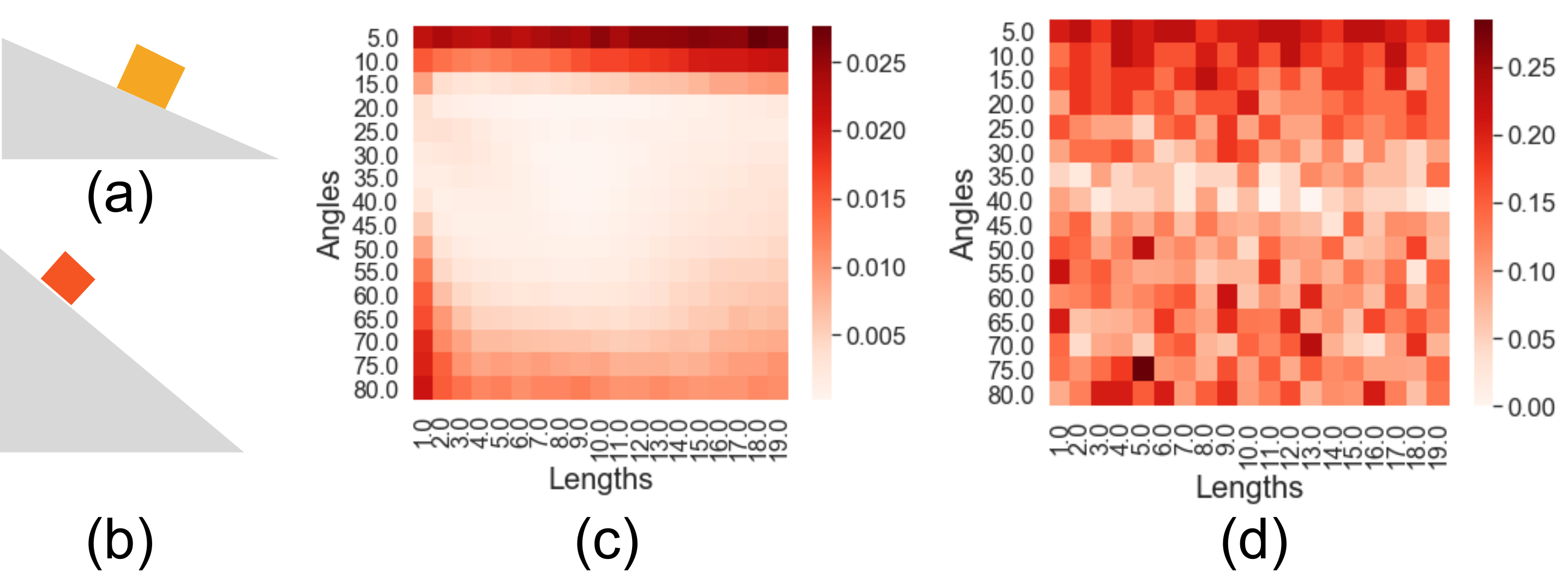}
\caption{(a,b) Examples of planes with two lengths and angles. MSE between true and inferred friction for 304 simulators (lighter is better) using (c) MetaVAE and (d) VAE.}
\label{fig:physics}
\end{figure}
Building on \cite{le2016inference}, we tackle this problem with ``meta-compiled inference" and optimize: 
\begin{equation}
    \mathcal{L}_{\phi} = \mathbb{E}_{p_{\theta_i^*} \sim p_{\mathcal{M}}} \mathbb{E}_{\vx \sim p_{\theta_i^*}(\vx)}\left[-g_\phi(\vz| p_{\theta_i^*}, \vx) \right]
\end{equation}
The meta-distribution $\mathcal{M}$ represents all possible simulators of planes with $L \in [1,20]$ and $A \in [5,85]$ degrees, and $p_{\theta_i^*}(\vx, \vz)$ represents a fixed simulator. The marginal distribution, $p_{\theta_i^*}(\vx)$ is obtained by repeatedly simulating to build a data set $\mathcal{D}_i = \{ \vx \}$. Thus the empirical meta-inference model $\hat{g}_\phi(\mathcal{D}_i, \vx)$ takes the data set and the output of a single simulation $\vx$ as input. We amortize over 25 simulators with $L \in \{2,4,6,8,10\}$ and $A \in \{20,30,40,50,60\}$, and model $\vz$ as a continuous 1-D random variable (interpreted as friction). After training the MetaVAE, we measure the mean squared error between the true and inferred friction for unseen simulators from $\mathcal{M}$.
Despite seeing only 25 out of 304 simulators, the MetaVAE transfers well: we get less than 0.001 MSE for $A \in [20,70]$ and $L \in [2,20]$. A standard VAE trained on a single simulator ($L=10$, $A=45$) exhibits both much worse generalization performance and greater error overall (notice the scale in the legends). 
\section{Demo: Learning Distribution Statistics}

Next, we explore whether the MetaVAE is capable of "meta-learning" the concept of a sufficient statistic for exponential families~\cite{wainwright2008graphical}.
Given a set of random samples, a sufficient statistic is a function that maps this set to a vector in $\mathbb{R}^d$. For the exponential families, where each family member has the form \(p(x) \propto \exp (\theta \cdot \phi(x))\) for some parameter \(\theta\), this vector can be used to estimate the parameters of the distribution.
In other words, the random samples (dataset) can be fully summarized by
the sufficient statistic, without any loss of information. 
Now consider a \textit{vector} of random variables $(x_1, \cdots, x_k)$, each distributed i.i.d from the same distribution with sufficient statistic $\phi(x_i)$. 
For exponential families, the sum $\sum_{i=1}^k \phi(x_i)$
is a sufficient statistic for the random vector. 
As an example, the number of successes is a sufficient statistic for a vector of i.i.d. Bernoulli, and the sample mean and variance are for a vector of Gaussians. 
With this intuition, we ask the following: having seen many realizations of random vectors from different exponential family distributions, can we learn a sufficient statistic for 
a new random vector that will be sufficient for estimating the parameters of its unseen, underlying distribution?
We aim to use the MetaVAE's meta-inference network to learn this mapping.
More precisely, the meta inference model $g_\phi(p_{\mathcal{D}_i},\vx)$ should act (as a function of $\vx$) as a sufficient statistic for an unseen distribution $p_{\mathcal{D}_i}$. 

\subsection{Data and Model Setup}
In this experiment, we use 
Gaussian (fixed variance), log-normal (fixed variance), exponential, symmetric beta, Laplace (fixed location), and Weibull (fixed scale) 
as exponential families. We then construct a set $\mathcal{M}_{\mathcal{I}}$ of 20-D 
vectors of random variables where each component is  i.i.d. distributed according to the same distribution. 
By construction, a random variable in this set will have only one free parameter, which can be found using the statistic learned by the meta-inference network.
We further restrict $\mathcal{M}_{\mathcal{I}}$ by bounding the free parameter to be within a range (e.g. Gaussians with mean between -5 and 5). 
After training, we measure how well we can infer the distributional parameters using the meta-inference model as a learned statistic for observations from unseen distributions. We compute the mean squared error (MSE) between the inferred and true parameters. We refer the reader to the appendix for more details.
\subsection{Experiment Results}
\begin{figure}[h!]
\centering     
\subfigure[Gaussian]{\includegraphics[width=0.34\linewidth]{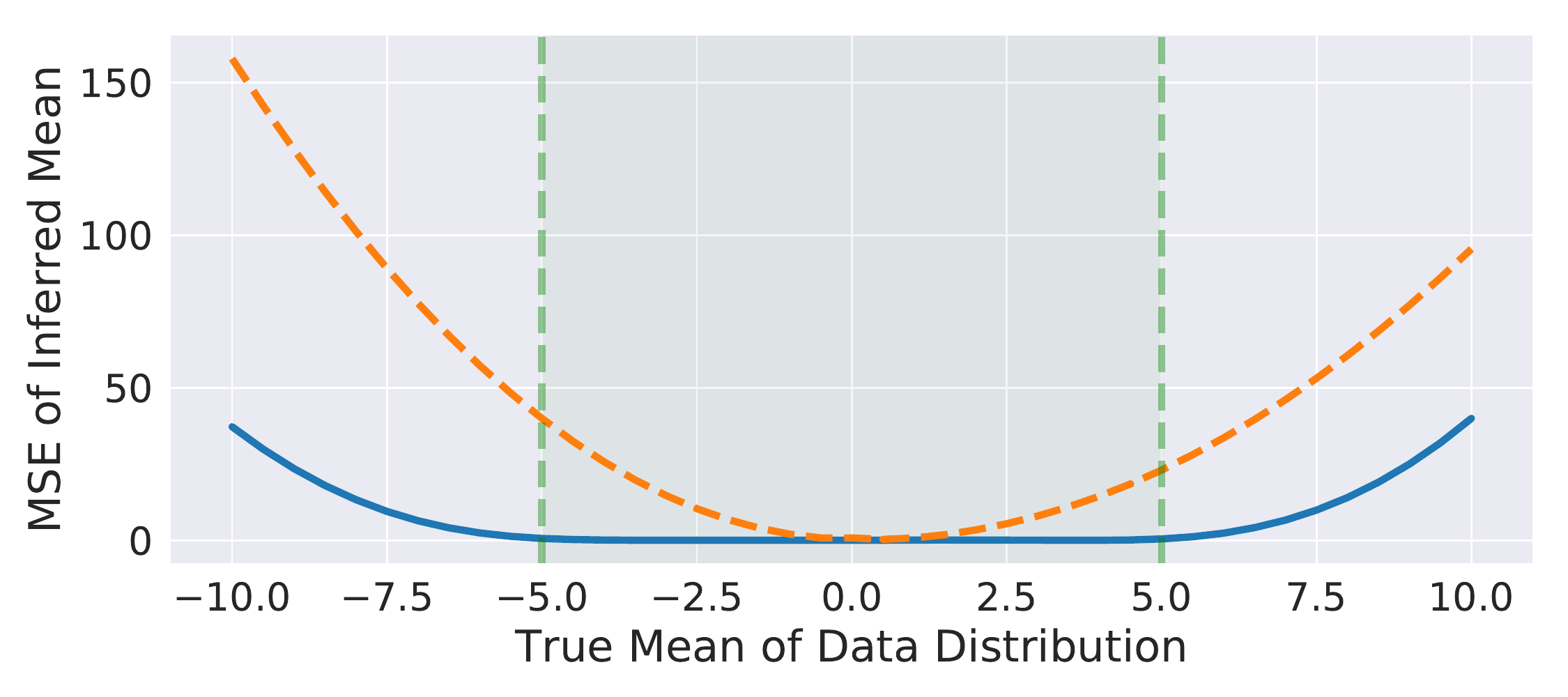}}
\subfigure[Log-Normal]
{\includegraphics[width=0.31\linewidth]{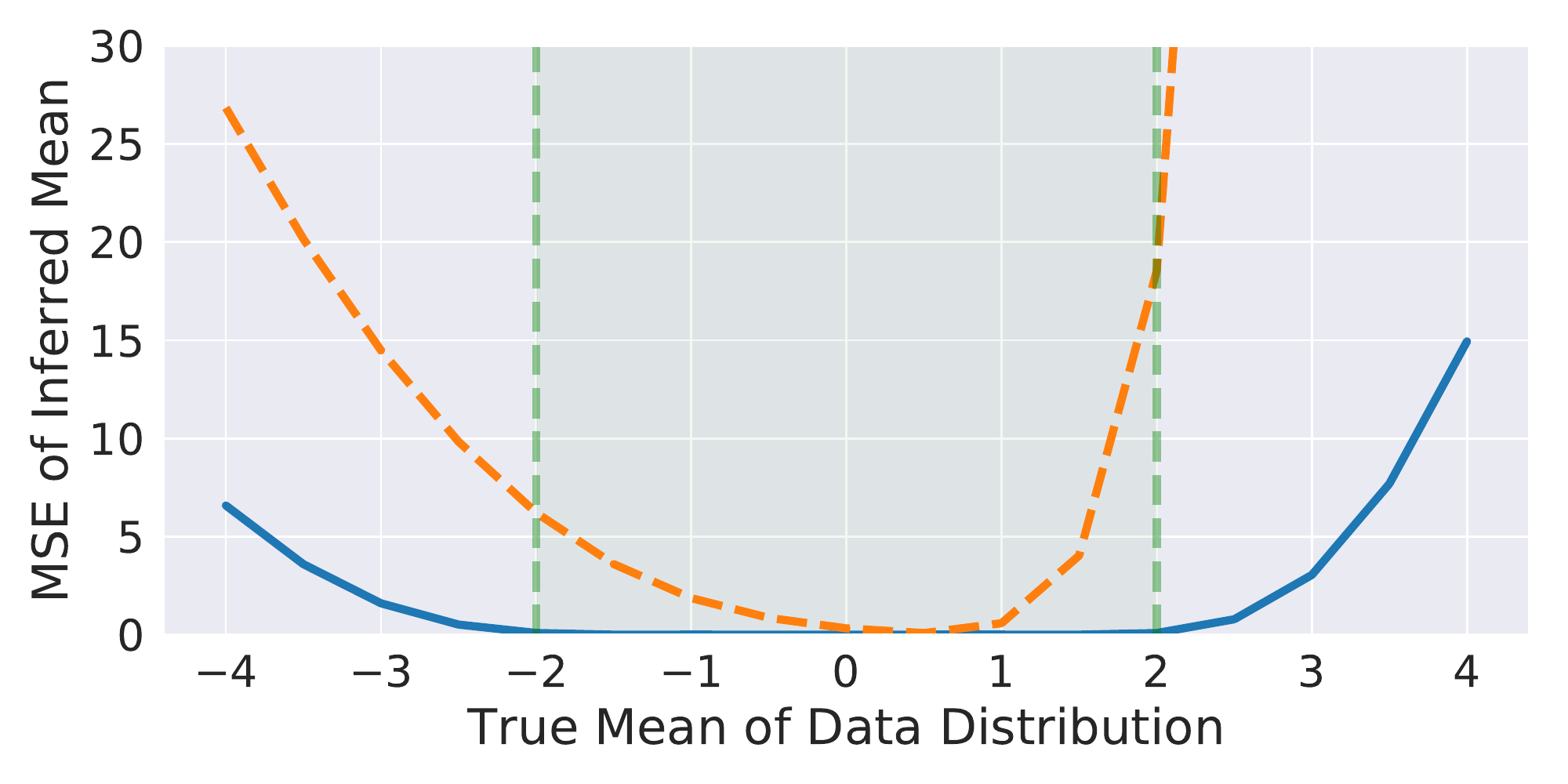}}
\subfigure[Exponential]
{\includegraphics[width=0.31\linewidth]{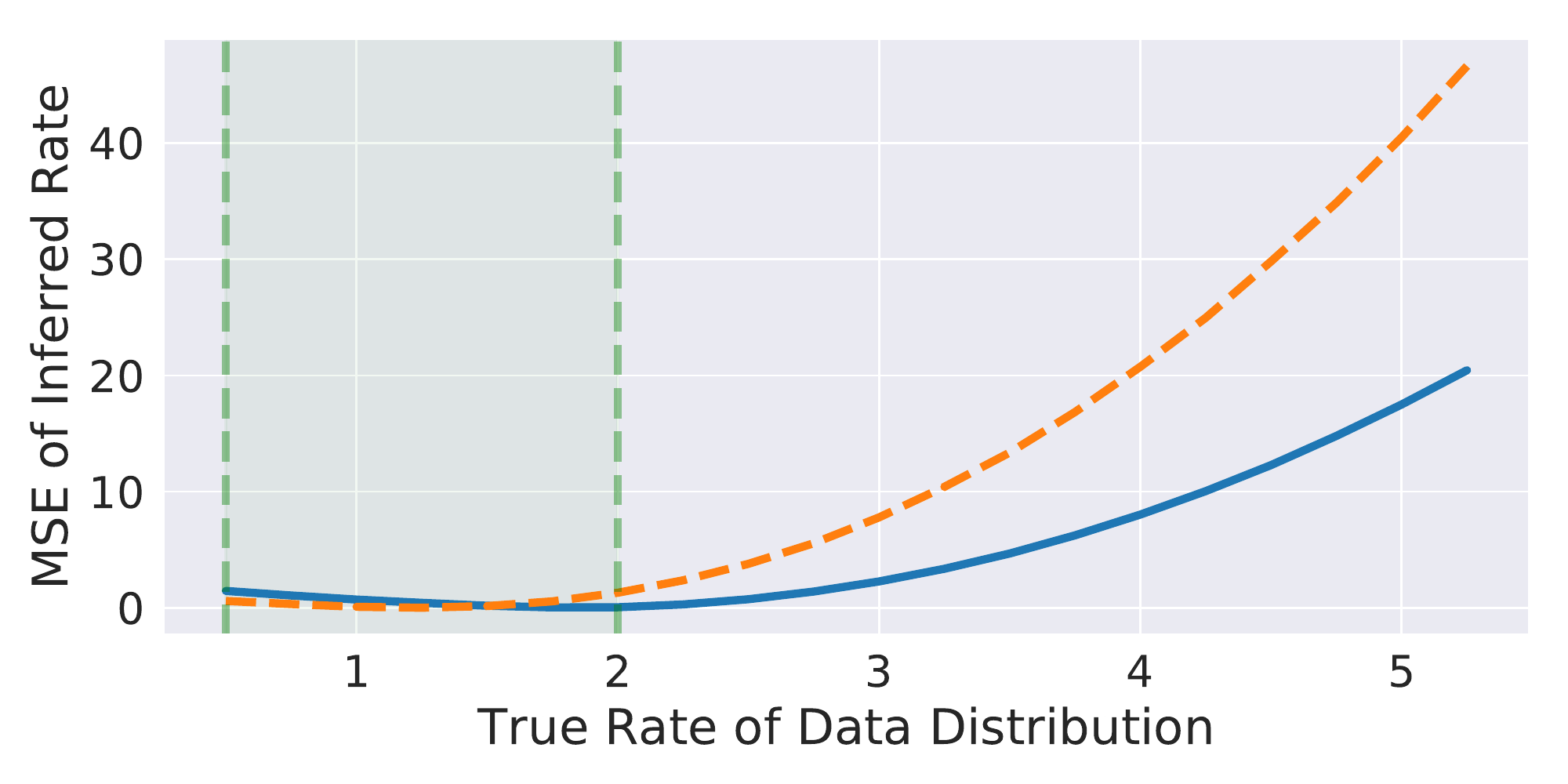}}
\caption{(a) MSE between the true and inferred mean as the true mean of $p_{\mathcal{D}_i}$ spans $[-10, 10]$. The green region shows the  meta-distribution. 
The orange (dashed) line shows a singly-amortized VAE trained on a single $p_{\mathcal{D}_i}(\vx)$ with mean $[-1.2, 1.1]$ (randomly chosen) and the blue (solid) line shows the MetaVAE. 
(b,c) show the MSE between the true and inferred parameters. The orange line is a singly-amortized VAE trained on a  randomly chosen distribution ($[-0.5, 1.8]$ for log-normal; $[1.4, 2.8]$ for exponential).}
\label{fig:explognorm_plot}
\end{figure}

\paragraph{Single Exponential Family}
Each $p_{\mathcal{D}_i}(\vx) \in \mathcal{M}$ is Gaussian with a mean sampled from $U(-5, 5)$.  At test time, we measure inference quality on (1) new random vectors from $\mathcal{M}$ whose entries are distributed as Gaussians with unseen means sampled from $U(-5, 5)$, and (2) a larger meta-distribution by sampling means from $U(-20, 20)$. We find the MetaVAE successfully learns the mean of the underlying Gaussians. Interestingly, in Fig.~\ref{fig:gaussian_plot}(a), we find that the inference quality only decays near the boundary of the meta-distribution.
We compare the MetaVAE to a VAE trained on one Gaussian distribution and find that doubly-amortizing increases the inference quality dramatically.
Then we move to two new exponential families: we similarly construct 30 log-normal random vectors with means from $U(-2, 2)$ and 30 Exponential random vectors with rates sampled from $U(0, 3)$. Like above, Fig.~\ref{fig:explognorm_plot} shows good performance of meta-inference over $\mathcal{M}$ in each case.
\paragraph{Many Exponential Families} 
Finally, we amortize over many types of distributional families simultaneously: we construct sets of 30 Gaussian, 30 log-normal, and 30 exponential random vectors (same bounds as above) to train a MetaVAE. This setup raises an interesting question: can we do inference for new random vectors comprised of \textit{unseen members of the exponential family} (e.g. Weibull)? 

\begin{figure}[h]
\centering     
\subfigure{\includegraphics[ width=0.9\linewidth]{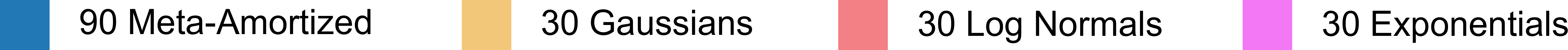}}
\subfigure[Gaussian]{\includegraphics[ width=0.32\linewidth]{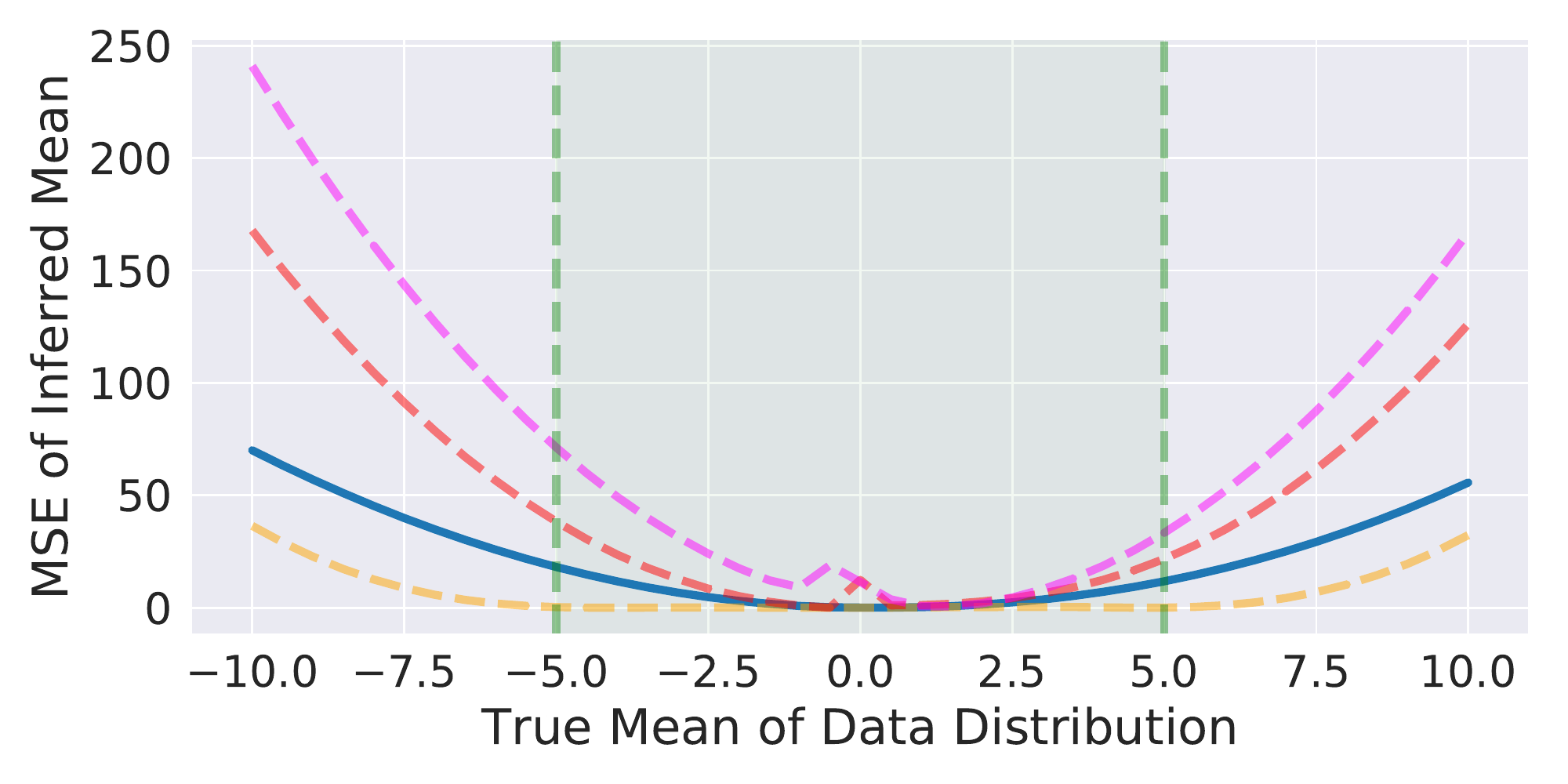}}
\subfigure[Log Normal]{\includegraphics[ width=0.32\linewidth]{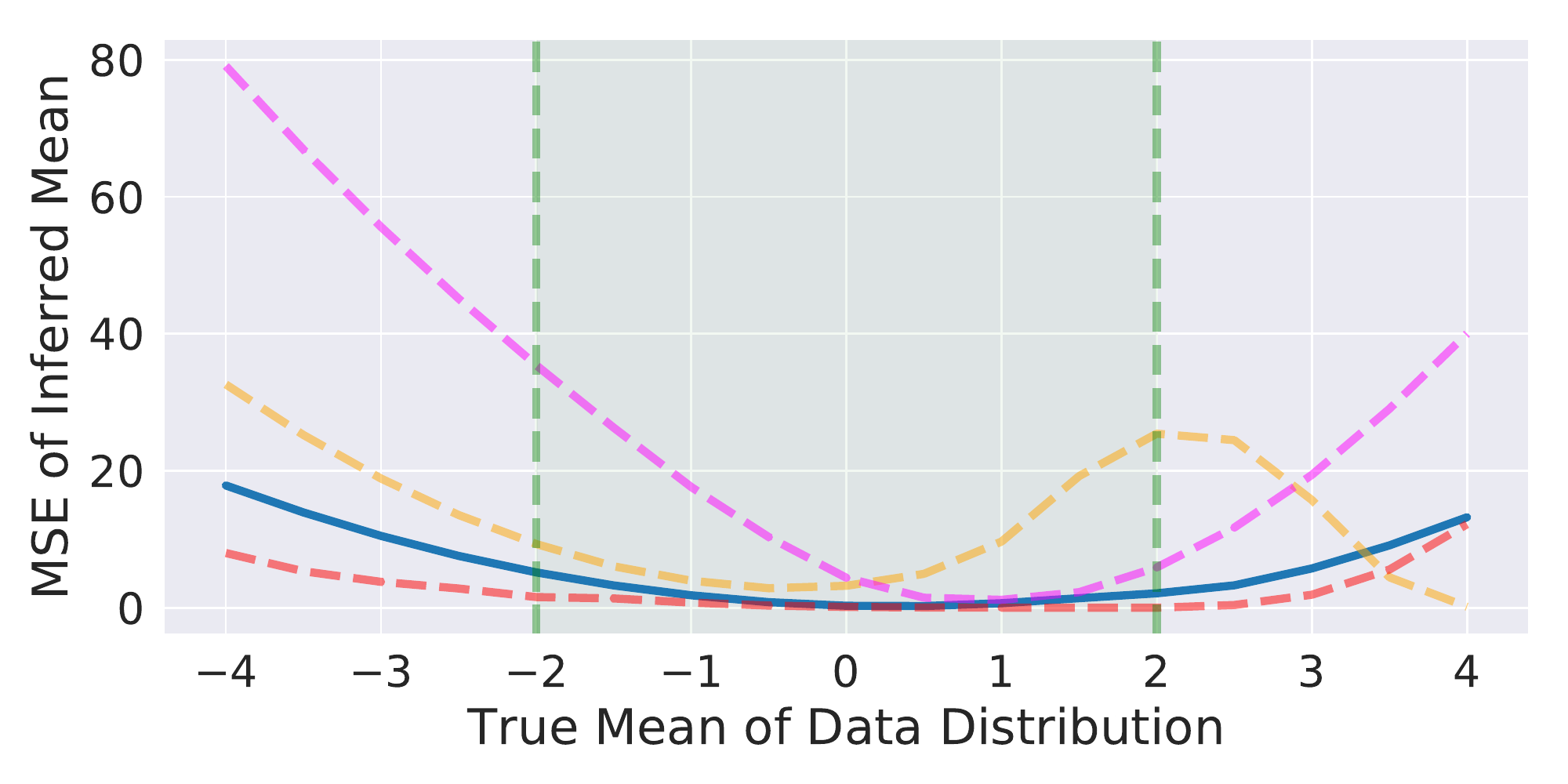}}
\subfigure[Exponential]{\includegraphics[ width=0.32\linewidth]{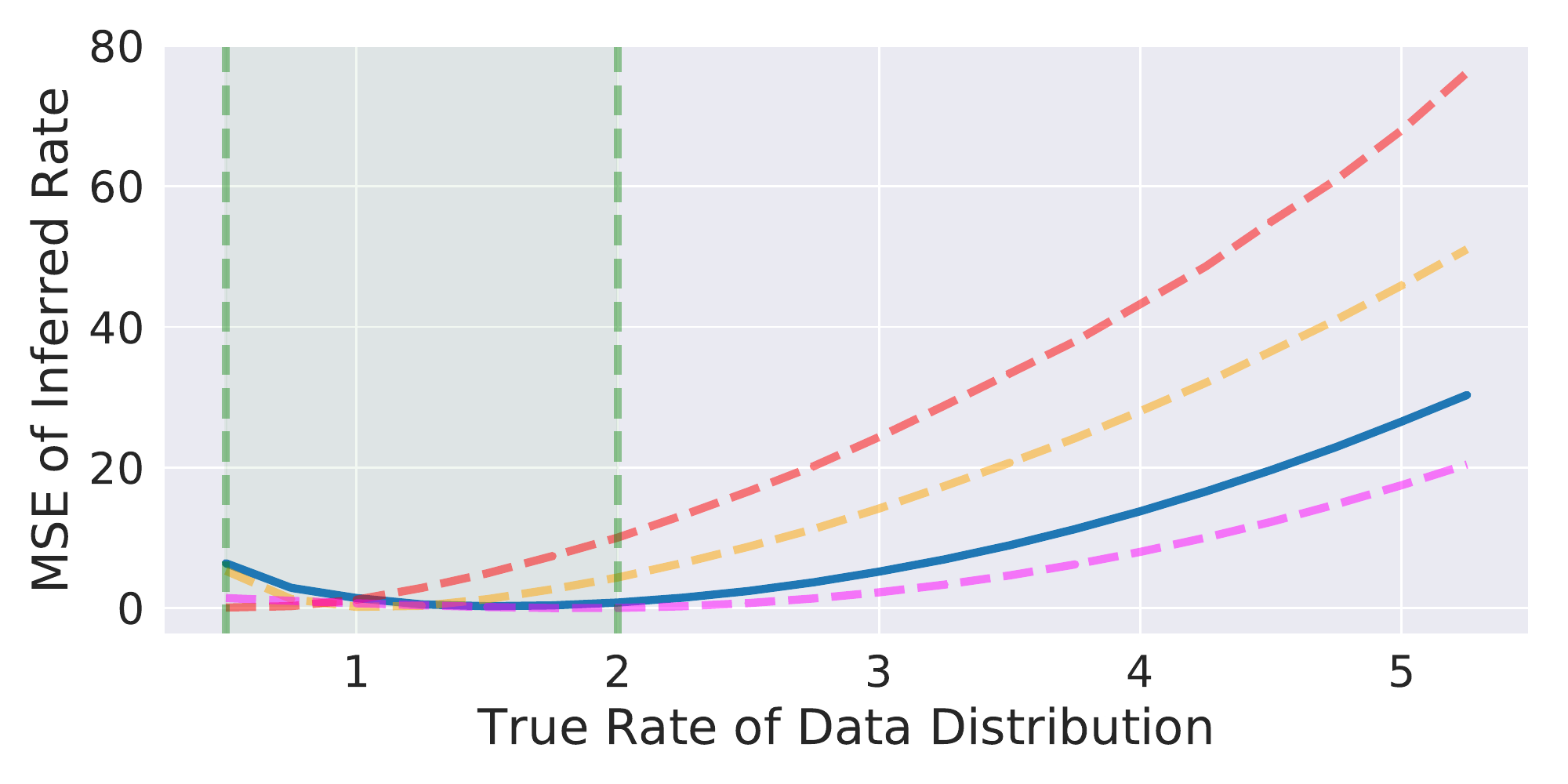}}
\subfigure[Beta($\alpha$, $\alpha$)]{\includegraphics[ width=0.32\linewidth]{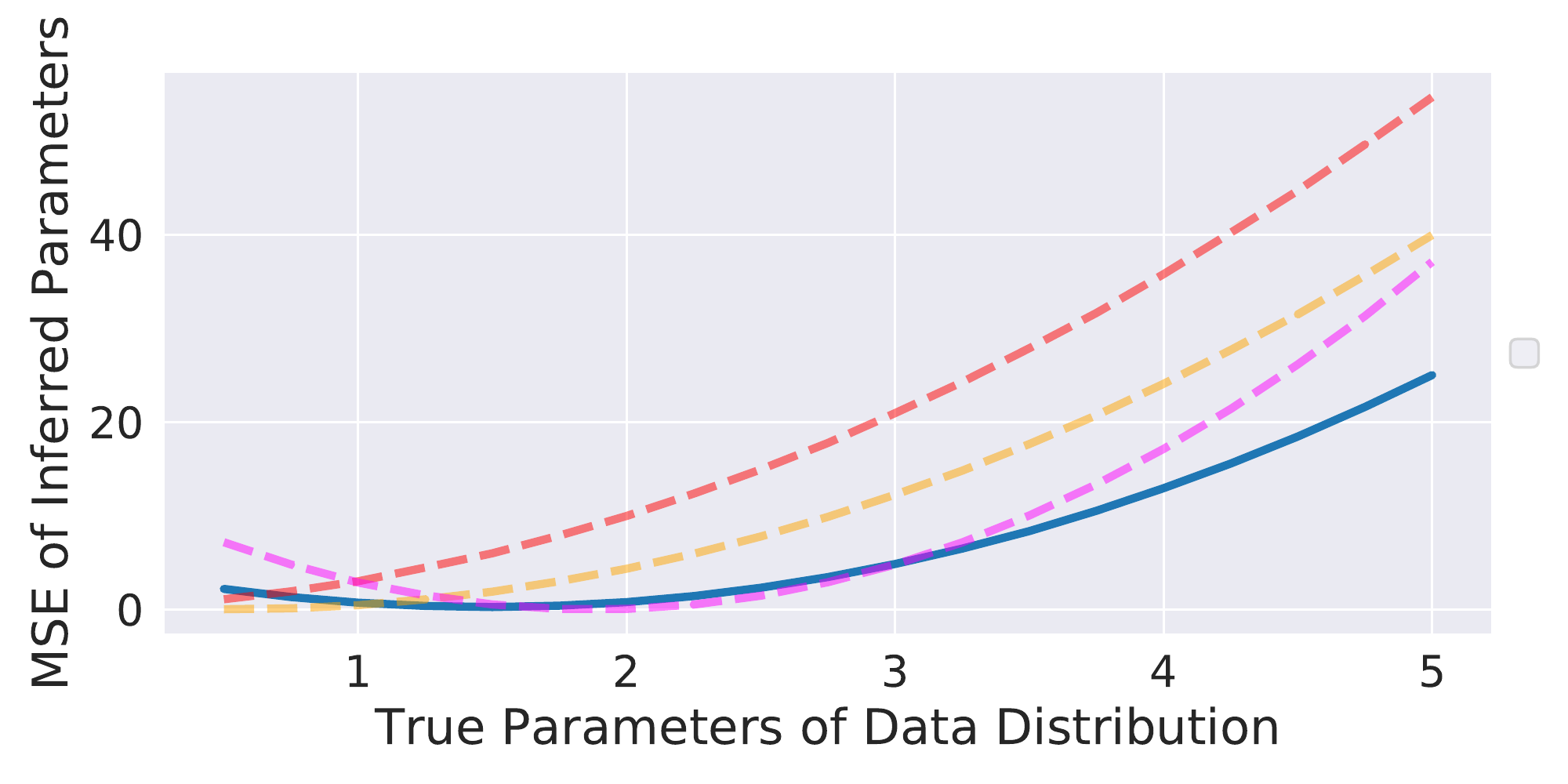}}
\subfigure[Weibull(scale=1)]{\includegraphics[ width=0.32\linewidth]{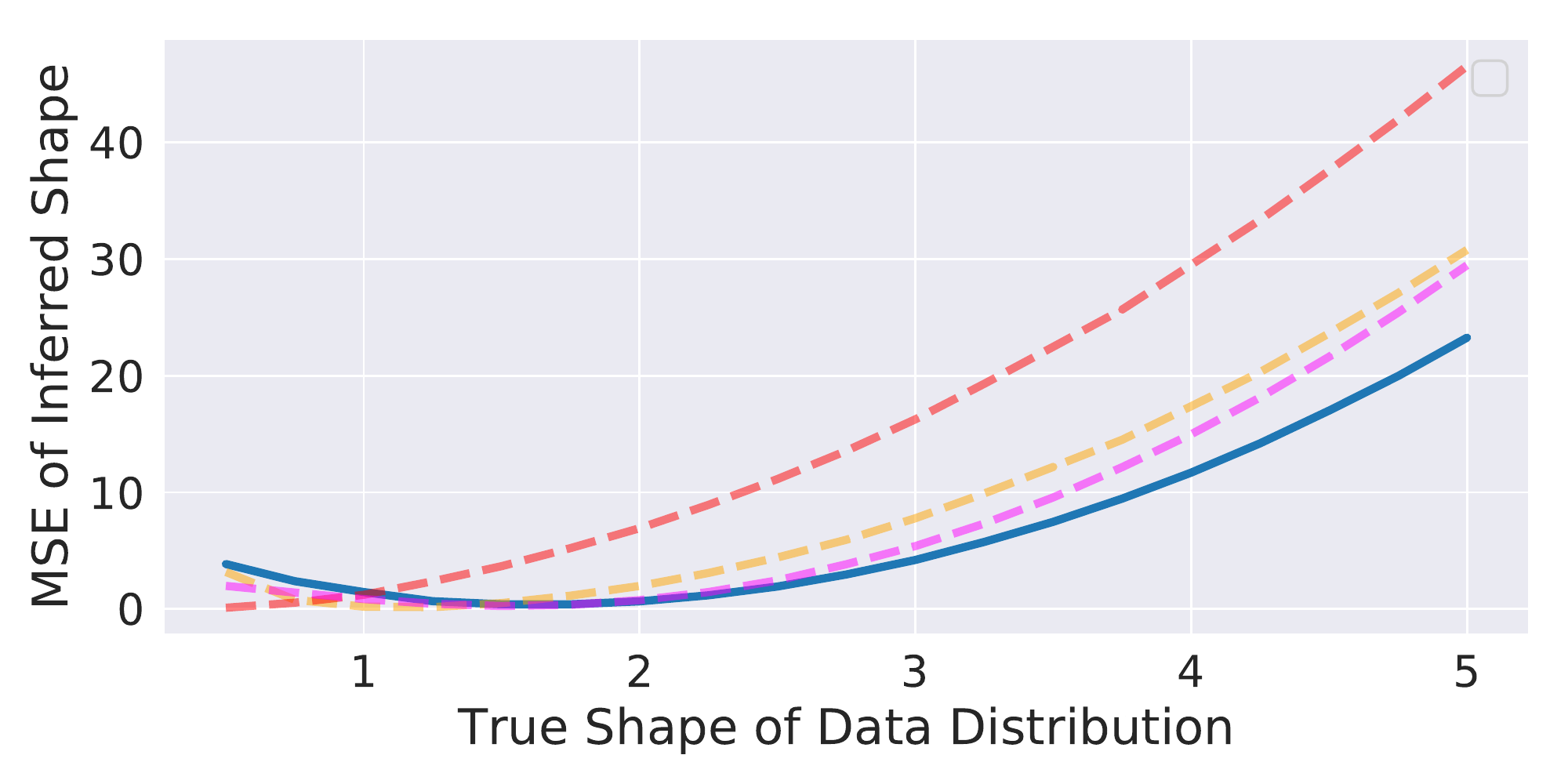}}
\subfigure[Laplace(loc=0)]{\includegraphics[ width=0.32\linewidth]{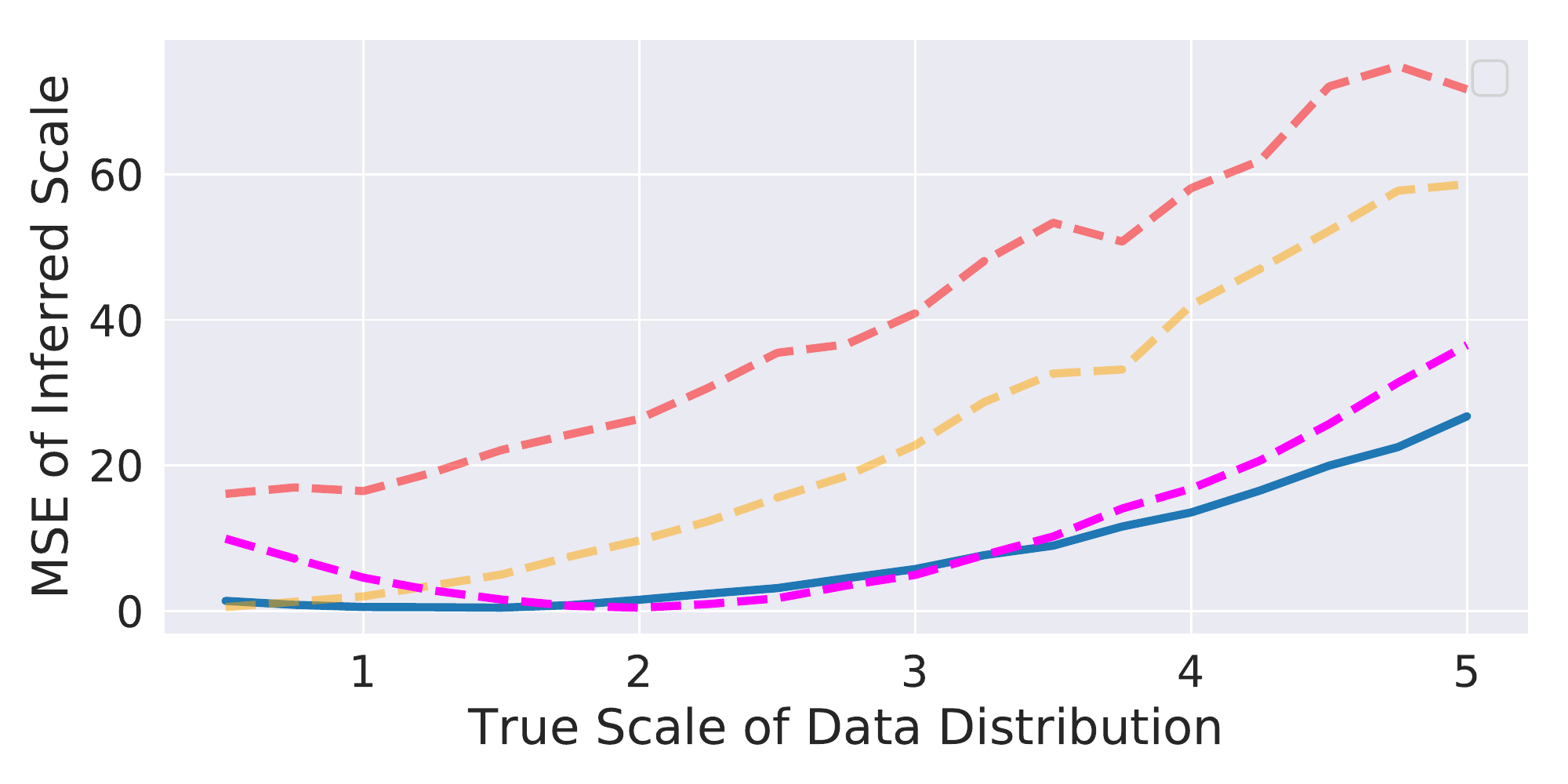}}
\caption{Comparison of a MetaVAE amortized over three members of the exponential family to MetaVAEs amortized over only a single member. Each subplot shows an unseen distribution from either the meta-distribution (b,c,d) or another exponential family (e,f,g).}
\label{fig:expfam_plot}
\end{figure}

We compare the performance a MetaVAE amortized over the 90 random vectors to 3 different (baseline) MetaVAEs, each of which is amortized over only 30 random vectors from one family (e.g. Gaussian). Below, Fig.~\ref{fig:expfam_plot}(a-c) plot the MSE of inferred and true parameters for Gaussian, log-normal, and exponential (all of which are in $\mathcal{M}$). Due to the double-amortization gap, the best performing model is the MetaVAE amortized on random vectors only from that family. However, the 90-amortized MetaVAE only performs slightly worse, beating the remaining two baselines dramatically. Next, Fig.~\ref{fig:expfam_plot}(d-f) show MSEs for three distributions not in $\mathcal{M}$: Weibull, Laplace, and Beta. The 90-amortized MetaVAE consistently outperforms all baselines. 

\section{Transformation-Invariance Experiments}
To motivate the next set of experiments, imagine designing a scene understanding algorithm for a self-driving car. The video datasets used to train deep learning agents are typically collected in isolated settings, such as in large cities during favorable weather conditions.
However, an agent deployed in the real world may face a variety of new settings such as paved roads in poorly-lit suburban areas.
In such cases, we would hope the agent could abstract away unnecessary sources of variation, such as different lighting conditions, and act upon more salient characteristics in the scene (e.g. pedestrians) that it has seen previously during training. 
Inference in this scenario would mean learning representations that are "transferable," or invariant to nuisance transformations such as time of day.
We take a step towards this goal as 
we study the MetaVAE for image distributions with explicit transformations, such as rotations or lighting. 
 
 \begin{figure}[h!]
    \centering
    \subfigure[Interleaved]{\includegraphics[ width=0.27\linewidth]{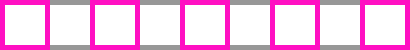}}
    \hspace{1em}
    \subfigure[Sparse]{\includegraphics[ width=0.27\linewidth]{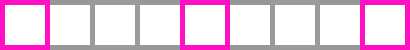}}
    \hspace{1em}
    \subfigure[Contiguous]{\includegraphics[ width=0.27\linewidth]{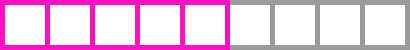}}
    \subfigure[Meta-Inference Pipeline]{\includegraphics[width=\linewidth]{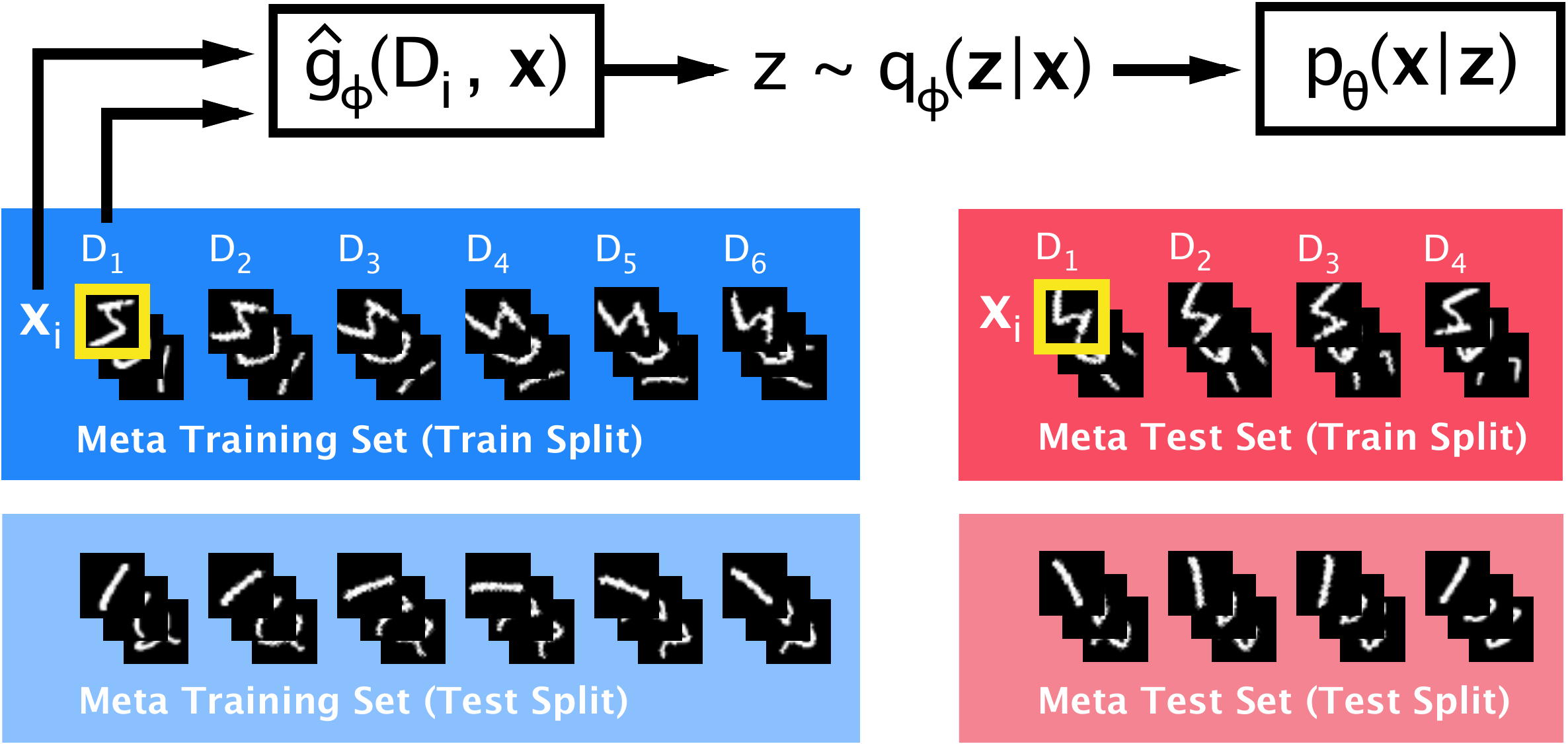}}
    \caption{(a-c) Three ways of defining the meta-training and meta-test splits; (b,c) pose a more difficult generalization challenge. (d) Overview of the doubly-amortized inference procedure. The meta-training set is used to train the MetaVAE (the test portion is to used to choose best parameters). The meta-test set is for evaluating the learned features, where the training portion is used to fit a linear classifier and the test portion is used to compute accuracy.}
    \label{fig:splits}
\end{figure}
 
\paragraph{Datasets} We study MNIST \cite{lecun1998mnist} and NORB \cite{lecun2004learning}, where we amortize over three axes of variation each (e.g. a range of camera angles or background lighting). Further, we vary how different variations are split into meta-training and meta-test sets, summarized in Fig.~\ref{fig:splits}(a-c). For instance, we may train the MetaVAE only on images with bright backgrounds and evaluate on darker images. We consider three meta-splits: \textit{interleaved}, where every other value in the range of possible transformations is selected; \textit{sparse}, where half the number of values are chosen as in interleaved; \textit{contiguous}, where we split the range in two ``contiguous" halves and train only over the first half. Each meta-split is a different measure of transfer-ability. 

\begin{figure*}[t!]
    \centering
    \includegraphics[width=0.9\textwidth]{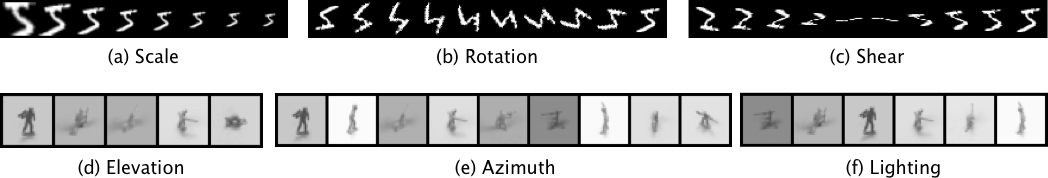}
    \caption{Examples of interpolating across three transformations each for MNIST and Small NORB. Notice that for NORB (unlike MNIST), other transformations are not held constant as we vary an individual axis.}
    \label{fig:datasets}
\end{figure*}
\begin{figure*}[t!]
    \centering
    \includegraphics[width=0.9\textwidth]{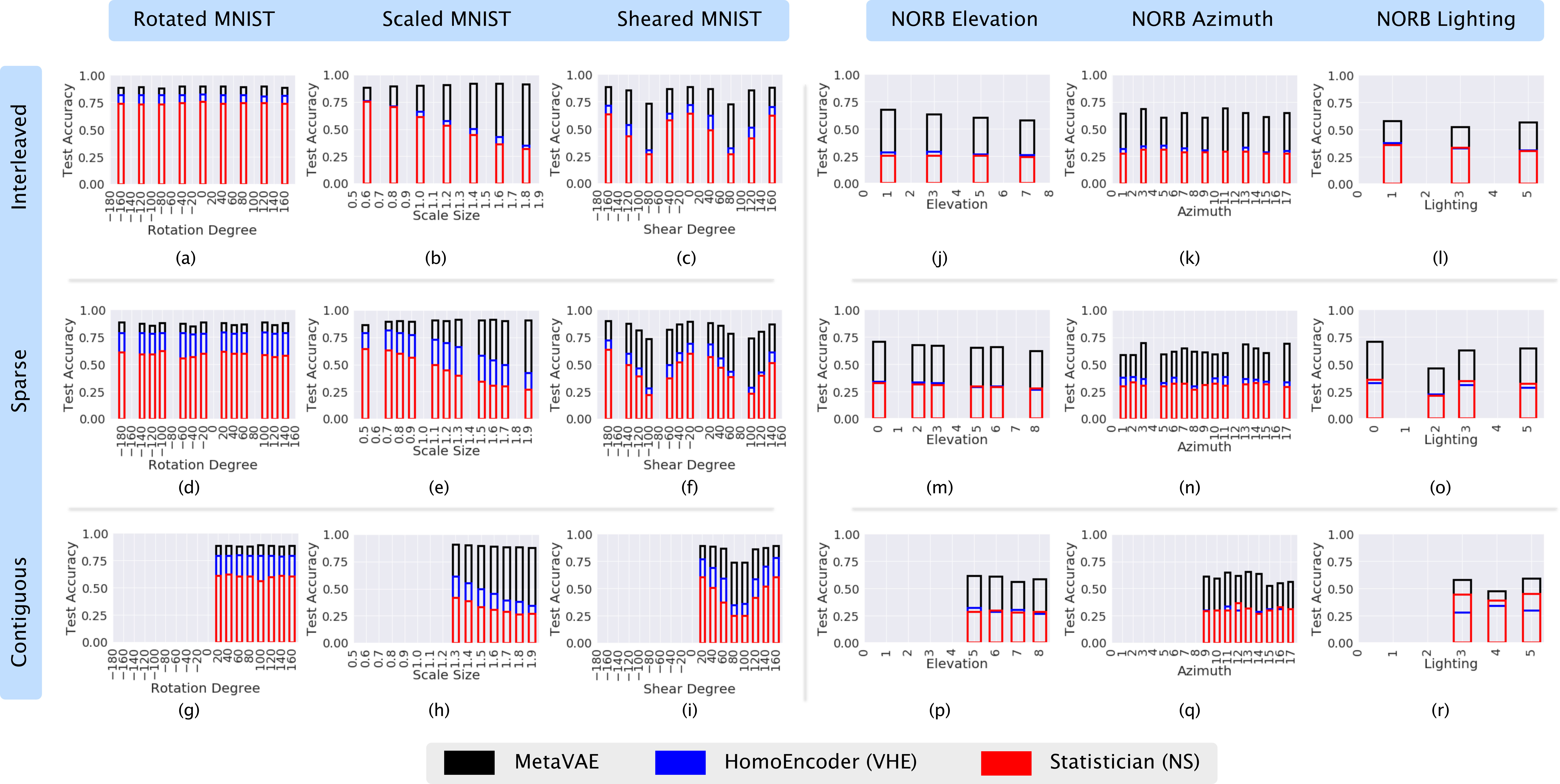}
    \caption{Classification Accuracy on Transformed MNIST and Small NORB for three different splits: interleaved, sparse, and contiguous. Each subfigure shows the prediction accuracy on the test set of held out transformations --- gaps represent the values used in training the amortized generative model. We compare the performance of MetaVAE (\textbf{black}), the homoencoder (\textcolor{blue}{blue}) and the statistician (\textcolor{red}{red}) and find appealing results for our proposed model.} 
    \label{fig:invariance:results}
\end{figure*}


\paragraph{Evaluation Metric} We evaluate the latent representations on a downstream classification task. Having trained the empirical meta-inference model $\hat{g}_\phi(\mathcal{D}, \vx)$ using the meta-train set, we then embed observations from a distribution in the meta-test set. Each time we ``embed" a test observation $\vx$, we feed in a data set $\mathcal{D}$ of samples \emph{from the meta-test set}. This way we construct a data set of latent features.

This feature set is split into a training and test subset. For both MNIST and NORB, each image has a corresponding label (e.g. digit or object class). 
Using the training portion (darker red in Fig.~\ref{fig:splits}d) , we fit a logistic regression classifier on the representations to predict the labels and compute accuracy on the test subset (lighter red in Fig.~\ref{fig:splits}d). 
Critically, logistic regression seeks the best linear split between classes in the latent space. For it to achieve good accuracy, such a linear division must already exist.
Thus, we treat a higher classification accuracy as a more transferable, invariant representation, as in \cite{berthelot2018understanding}.

\paragraph{Baselines} We compare the performance of MetaVAE against two baselines: the Neural Statistician (NS), a hierarchical VAE which models sets of observations with a global latent variable; and the Variational HomoEncoder (VHE), a more computationally-efficient variant of  NS. To ensure a fair comparison, we use the same hyperparameters and architectures across all models. See Appendix for details.  

\subsection{Transformed MNIST}

\paragraph{Dataset Construction} We artificially impose three axes of variations on MNIST digits. We transform each image with 18 \textit{rotations} (-180 to 180 by 20 degrees), 15 \textit{scales} (50\% to 200\% original size by 10\%), and  18 \textit{skews} (-180 to 180 by 20 degrees). 
See Fig.~\ref{fig:datasets}(a-c) for an example for a single digit. For each axes of variation, the other two are held constant e.g. skew and size are constant when varying rotation. 

\paragraph{Results} We find consistent evidence that MetaVAE features outperform both VHE and NS features across all settings, often by a significant margin. In particular, VHE and NS have decaying performance as scale increases to 2.0. Similarly, for extreme shear values near -80 and 80 degrees where the image is nearly flat (see Fig.~\ref{fig:datasets}c), VHE and NS again suffer greatly in performance.
However, MetaVAE features transfer better: we do not notice a drop in accuracy as scale increases and the effect of significant shearing is more gradual. 
This suggests that MetaVAE has learned some invariances to transformations that NS and VHE lack.

\subsection{Small NORB}


\paragraph{Dataset Construction} The NORB dataset contains grayscale images of real world toys belonging to five classes: animals, humans, airplanes, trucks, and cars. The objects were imaged under 6 \textit{lighting} conditions, 9 \textit{elevations} (30 to 70 degrees every 5 degrees), and 18 \textit{azimuths} (0 to 340 every 20 degrees). Unlike the MNIST dataset, extraneous transformations are \textit{not} held constant as one transformation is varied. For example, as Fig.~\ref{fig:datasets}(f) shows, the azimuth and elevation (randomly) change as we vary lighting. This design, while more difficult to amortize, is more realistic in real world datasets where it is too expensive to collect data holding all other variables constant.

\paragraph{Results}
The MetaVAE representations outperform those of VHE and NS by 10 to 35\% accuracy. Overall, we notice accuracies are much lower in NORB than in MNIST, which is likely due to the complexity of learning real world image distributions and randomness introduced by variations in extraneous transformations. 
We note that the strong performance of the MetaVAE despite varying transformations is promising support for our approach to meta-amortization, suggesting that the MetaVAE is able to ignore irrelevant signals while capturing the principal axes of variation.
\subsection{Analysis}
We aim to quantitatively measure the intuition that amortizing over a family of transformations should yield representations that are invariant to that transformation.
For example, how much does the representation change as we alter the rotation in MNIST from -180 to 180, or interpolate the background from dark to light in NORB? 

To investigate, we use a MetaVAE amortized over a family of transformations (e.g. interleaved rotations) and compare the average
L$_2$ distance between the learned representation of a base (default) image and those of every rotated image. As a baseline, we compare this distance to the average L$_2$ distance of a separate family of transformations (e.g. scale) that this MetaVAE was not amortized over (e.g. having only seen different rotations during training). Table~\ref{table:distance} shows the distances for MNIST and NORB. Consistently, the lowest distances belong to the class of transformations that the MetaVAE was amortized over, which supports the intuition about learning invariances.

\begin{table}[h]
\centering
\begin{tabular}{r|c|c|c}
\toprule
Model Dataset & Rotation & Scale & Skew \\
\midrule
Rotated MNIST & $\mathbf{1.65}$ & $4.44$ & $4.09$ \\ 
Scaled MNIST &  $5.44$ & $\mathbf{2.16}$ & $4.92$ \\  
Skewed MNIST & $3.79$ & $4.89$ & $\mathbf{1.47}$ \\   
\toprule
Model Dataset & Elevation & Azimuth & Lighting \\
\midrule
NORB Elevation & $\mathbf{0.39}$ & $1.16$ & $1.27$ \\ 
NORB Azimuth &  $1.42$ & $\mathbf{0.44}$ & $1.26$ \\  
NORB Lighting & $1.69$ & $1.27$ & $\mathbf{0.26}$ \\
\bottomrule
\end{tabular}
\caption{L$_2$ distances between MetaVAE representations. Each row indicates the datasets  used for training; each column indicates the datasets used to compute representations.}
\label{table:distance}
\end{table}
\section{Conclusion}
In summary, we considered constructing an algorithm that can do inference for a  \textit{family} of probabilistic models. We introduced a meta-amortized inference paradigm and a new generative model, the MetaVAE. 
Through experiments on MNIST and Small NORB, we showed that the MetaVAE learned transferable representations that generalize well across similar data distributions in downstream tasks. 
Future work could consider new applications of meta-inference in video prediction \cite{ramanathan2015learning}.

\bibliographystyle{aaai}
\bibliography{meta}

\begin{thebibliography}{}

\bibitem[\protect\citeauthoryear{Bartunov and Vetrov}{2016}]{bartunov2016fast}
Bartunov, S., and Vetrov, D.~P.
\newblock 2016.
\newblock Fast adaptation in generative models with generative matching
  networks.
\newblock {\em arXiv preprint arXiv:1612.02192}.

\bibitem[\protect\citeauthoryear{Berthelot \bgroup et al\mbox.\egroup
  }{2018}]{berthelot2018understanding}
Berthelot, D.; Raffel, C.; Roy, A.; and Goodfellow, I.
\newblock 2018.
\newblock Understanding and improving interpolation in autoencoders via an
  adversarial regularizer.
\newblock {\em arXiv preprint arXiv:1807.07543}.

\bibitem[\protect\citeauthoryear{Blei, Kucukelbir, and
  McAuliffe}{2017}]{blei2017variational}
Blei, D.~M.; Kucukelbir, A.; and McAuliffe, J.~D.
\newblock 2017.
\newblock Variational inference: A review for statisticians.
\newblock {\em Journal of the American Statistical Association}
  112(518):859--877.

\bibitem[\protect\citeauthoryear{Bornschein and
  Bengio}{2014}]{bornschein2014reweighted}
Bornschein, J., and Bengio, Y.
\newblock 2014.
\newblock Reweighted wake-sleep.
\newblock {\em arXiv preprint arXiv:1406.2751}.

\bibitem[\protect\citeauthoryear{Brock, Donahue, and
  Simonyan}{2018}]{brock2018large}
Brock, A.; Donahue, J.; and Simonyan, K.
\newblock 2018.
\newblock Large scale gan training for high fidelity natural image synthesis.
\newblock {\em arXiv preprint arXiv:1809.11096}.

\bibitem[\protect\citeauthoryear{Edwards and
  Storkey}{2016}]{edwards2016towards}
Edwards, H., and Storkey, A.
\newblock 2016.
\newblock Towards a neural statistician.
\newblock {\em arXiv preprint arXiv:1606.02185}.

\bibitem[\protect\citeauthoryear{Finn, Abbeel, and
  Levine}{2017}]{finn2017model}
Finn, C.; Abbeel, P.; and Levine, S.
\newblock 2017.
\newblock Model-agnostic meta-learning for fast adaptation of deep networks.
\newblock In {\em Proceedings of the 34th International Conference on Machine
  Learning-Volume 70},  1126--1135.
\newblock JMLR. org.

\bibitem[\protect\citeauthoryear{Garnelo \bgroup et al\mbox.\egroup
  }{2018}]{garnelo2018neural}
Garnelo, M.; Schwarz, J.; Rosenbaum, D.; Viola, F.; Rezende, D.~J.; Eslami, S.;
  and Teh, Y.~W.
\newblock 2018.
\newblock Neural processes.
\newblock {\em arXiv preprint arXiv:1807.01622}.

\bibitem[\protect\citeauthoryear{Gelfand and Smith}{1990}]{gelfand1990sampling}
Gelfand, A.~E., and Smith, A.~F.
\newblock 1990.
\newblock Sampling-based approaches to calculating marginal densities.
\newblock {\em Journal of the American statistical association}
  85(410):398--409.

\bibitem[\protect\citeauthoryear{Gershman and
  Goodman}{2014}]{gershman2014amortized}
Gershman, S., and Goodman, N.
\newblock 2014.
\newblock Amortized inference in probabilistic reasoning.
\newblock In {\em Proceedings of the Annual Meeting of the Cognitive Science
  Society}, volume~36.

\bibitem[\protect\citeauthoryear{Gordon \bgroup et al\mbox.\egroup
  }{2018}]{gordon2018decision}
Gordon, J.; Bronskill, J.; Bauer, M.; Nowozin, S.; and Turner, R.~E.
\newblock 2018.
\newblock Decision-theoretic meta-learning: Versatile and efficient
  amortization of few-shot learning.
\newblock {\em arXiv preprint arXiv:1805.09921}.

\bibitem[\protect\citeauthoryear{Grant \bgroup et al\mbox.\egroup
  }{2018}]{grant2018recasting}
Grant, E.; Finn, C.; Levine, S.; Darrell, T.; and Griffiths, T.
\newblock 2018.
\newblock Recasting gradient-based meta-learning as hierarchical bayes.
\newblock {\em arXiv preprint arXiv:1801.08930}.

\bibitem[\protect\citeauthoryear{Hastings}{1970}]{hastings1970monte}
Hastings, W.~K.
\newblock 1970.
\newblock Monte carlo sampling methods using markov chains and their
  applications.

\bibitem[\protect\citeauthoryear{Hewitt \bgroup et al\mbox.\egroup
  }{2018}]{hewitt2018variational}
Hewitt, L.~B.; Nye, M.~I.; Gane, A.; Jaakkola, T.; and Tenenbaum, J.~B.
\newblock 2018.
\newblock The variational homoencoder: Learning to learn high capacity
  generative models from few examples.
\newblock {\em arXiv preprint arXiv:1807.08919}.

\bibitem[\protect\citeauthoryear{Hinton \bgroup et al\mbox.\egroup
  }{1995}]{hinton1995wake}
Hinton, G.~E.; Dayan, P.; Frey, B.~J.; and Neal, R.~M.
\newblock 1995.
\newblock The" wake-sleep" algorithm for unsupervised neural networks.
\newblock {\em Science} 268(5214):1158--1161.

\bibitem[\protect\citeauthoryear{Jordan \bgroup et al\mbox.\egroup
  }{1999}]{jordan1999introduction}
Jordan, M.~I.; Ghahramani, Z.; Jaakkola, T.~S.; and Saul, L.~K.
\newblock 1999.
\newblock An introduction to variational methods for graphical models.
\newblock {\em Machine learning} 37(2):183--233.

\bibitem[\protect\citeauthoryear{Kim \bgroup et al\mbox.\egroup
  }{2019}]{kim2019attentive}
Kim, H.; Mnih, A.; Schwarz, J.; Garnelo, M.; Eslami, A.; Rosenbaum, D.;
  Vinyals, O.; and Teh, Y.~W.
\newblock 2019.
\newblock Attentive neural processes.
\newblock {\em arXiv preprint arXiv:1901.05761}.

\bibitem[\protect\citeauthoryear{Kingma and Welling}{2013}]{kingma2013auto}
Kingma, D.~P., and Welling, M.
\newblock 2013.
\newblock Auto-encoding variational bayes.
\newblock {\em arXiv preprint arXiv:1312.6114}.

\bibitem[\protect\citeauthoryear{Klingler \bgroup et al\mbox.\egroup
  }{2017}]{klingler2017efficient}
Klingler, S.; Wampfler, R.; K{\"a}ser, T.; Solenthaler, B.; and Gross, M.~H.
\newblock 2017.
\newblock Efficient feature embeddings for student classification with
  variational auto-encoders.
\newblock In {\em EDM}.

\bibitem[\protect\citeauthoryear{Le, Baydin, and Wood}{2016}]{le2016inference}
Le, T.~A.; Baydin, A.~G.; and Wood, F.
\newblock 2016.
\newblock Inference compilation and universal probabilistic programming.
\newblock {\em arXiv preprint arXiv:1610.09900}.

\bibitem[\protect\citeauthoryear{Le \bgroup et al\mbox.\egroup
  }{2018}]{le2018revisiting}
Le, T.~A.; Kosiorek, A.~R.; Siddharth, N.; Teh, Y.~W.; and Wood, F.
\newblock 2018.
\newblock Revisiting reweighted wake-sleep.
\newblock {\em arXiv preprint arXiv:1805.10469}.

\bibitem[\protect\citeauthoryear{LeCun \bgroup et al\mbox.\egroup
  }{2004}]{lecun2004learning}
LeCun, Y.; Huang, F.~J.; Bottou, L.; et~al.
\newblock 2004.
\newblock Learning methods for generic object recognition with invariance to
  pose and lighting.
\newblock In {\em CVPR (2)},  97--104.
\newblock Citeseer.

\bibitem[\protect\citeauthoryear{LeCun}{1998}]{lecun1998mnist}
LeCun, Y.
\newblock 1998.
\newblock The mnist database of handwritten digits.
\newblock {\em http://yann. lecun. com/exdb/mnist/}.

\bibitem[\protect\citeauthoryear{Mao \bgroup et al\mbox.\egroup
  }{2018}]{mao2018deep}
Mao, C.; Yao, L.; Pan, Y.; Luo, Y.; and Zeng, Z.
\newblock 2018.
\newblock Deep generative classifiers for thoracic disease diagnosis with chest
  x-ray images.
\newblock In {\em 2018 IEEE International Conference on Bioinformatics and
  Biomedicine (BIBM)},  1209--1214.
\newblock IEEE.

\bibitem[\protect\citeauthoryear{Oord \bgroup et al\mbox.\egroup
  }{2016}]{oord2016wavenet}
Oord, A. v.~d.; Dieleman, S.; Zen, H.; Simonyan, K.; Vinyals, O.; Graves, A.;
  Kalchbrenner, N.; Senior, A.; and Kavukcuoglu, K.
\newblock 2016.
\newblock Wavenet: A generative model for raw audio.
\newblock {\em arXiv preprint arXiv:1609.03499}.

\bibitem[\protect\citeauthoryear{Ramanathan \bgroup et al\mbox.\egroup
  }{2015}]{ramanathan2015learning}
Ramanathan, V.; Tang, K.; Mori, G.; and Fei-Fei, L.
\newblock 2015.
\newblock Learning temporal embeddings for complex video analysis.
\newblock In {\em Proceedings of the IEEE International Conference on Computer
  Vision},  4471--4479.

\bibitem[\protect\citeauthoryear{Reed \bgroup et al\mbox.\egroup
  }{2017}]{reed2017few}
Reed, S.; Chen, Y.; Paine, T.; Oord, A. v.~d.; Eslami, S.; Rezende, D.;
  Vinyals, O.; and de~Freitas, N.
\newblock 2017.
\newblock Few-shot autoregressive density estimation: Towards learning to learn
  distributions.
\newblock {\em arXiv preprint arXiv:1710.10304}.

\bibitem[\protect\citeauthoryear{Segler \bgroup et al\mbox.\egroup
  }{2017}]{segler2017generating}
Segler, M.~H.; Kogej, T.; Tyrchan, C.; and Waller, M.~P.
\newblock 2017.
\newblock Generating focused molecule libraries for drug discovery with
  recurrent neural networks.
\newblock {\em ACS central science} 4(1):120--131.

\bibitem[\protect\citeauthoryear{Snell, Swersky, and
  Zemel}{2017}]{snell2017prototypical}
Snell, J.; Swersky, K.; and Zemel, R.
\newblock 2017.
\newblock Prototypical networks for few-shot learning.
\newblock In {\em Advances in Neural Information Processing Systems},
  4077--4087.

\bibitem[\protect\citeauthoryear{Vinyals \bgroup et al\mbox.\egroup
  }{2016}]{vinyals2016matching}
Vinyals, O.; Blundell, C.; Lillicrap, T.; Wierstra, D.; et~al.
\newblock 2016.
\newblock Matching networks for one shot learning.
\newblock In {\em Advances in Neural Information Processing Systems},
  3630--3638.

\bibitem[\protect\citeauthoryear{Wainwright and
  Jordan}{2008}]{wainwright2008graphical}
Wainwright, M.~J., and Jordan, M.~I.
\newblock 2008.
\newblock Graphical models, exponential families, and variational inference.
\newblock {\em Foundations and Trends{\textregistered} in Machine Learning}
  1(1--2):1--305.

\bibitem[\protect\citeauthoryear{Yildirim}{2014}]{yildirim2014perception}
Yildirim, I.
\newblock 2014.
\newblock From perception to conception: learning multisensory representations.

\end{thebibliography}

\appendix
\newpage

\section*{Appendix}
\section{Relationship to Bayesian Neural Networks}
The MetaVAE is closely related to a fully Bayesian VAE where one would explicitly model a posterior distribution over parameters. More precisely, this involves the factorization of the joint, $p(\vx,\vz,\theta) = p(\vx|\vz,\theta)p(\vz)p(\theta)$. Then, the appropriate inference network would be $q_\phi(\vz|\theta, \vx)$ i.e. amortized over a family of generative models $\{p(\vx,\vz,\theta), \theta \in \Theta\}$. If $\Theta$ is a finite set, then the fully Bayesian VAE is analogous to a MetaVAE. In practice, Bayesian neural networks are difficult to train. By discretizing $\Theta$ to a finite set, we make the problem tractable.

\section{Demo: Clustering Mixtures (Continued)}
We provide additional details for the experimental setup outlined in the main text. Formally, we let each distribution $p_{\mathcal{D}_i}(\vx) \sim p_\mathcal{M}$ be a MoG, where $p_{\mathcal{D}}(\vx) = \frac{1}{2}\mathcal{N}(\mathbf{\mu}_1, 0.1) + \frac{1}{2}\mathcal{N}(\mathbf{\mu}_2, 0.1)$. Each equally-mixed Gaussian component has isotropic covariance of 0.1 and mean drawn from $U(-5, 5)$. We assign each mixture component a label of 0 or 1. 
Therefore, we represent each $p_{\mathcal{D}_i}(\vx)$ as a data set of samples $\mathcal{D}_i = \{\vx_1, ..., \vx_N\} \sim p_{\mathcal{D}_i}(\vx)$ in our inference procedure. 
The meta-inference model $\hat{g}_\phi(\mathcal{D}_i, \vx)$ takes as input the data set as well as an observation $\vx \sim  p_{\mathcal{D}_i}(\vx)$.

\begin{figure}[h!]
\includegraphics[ width=\linewidth]{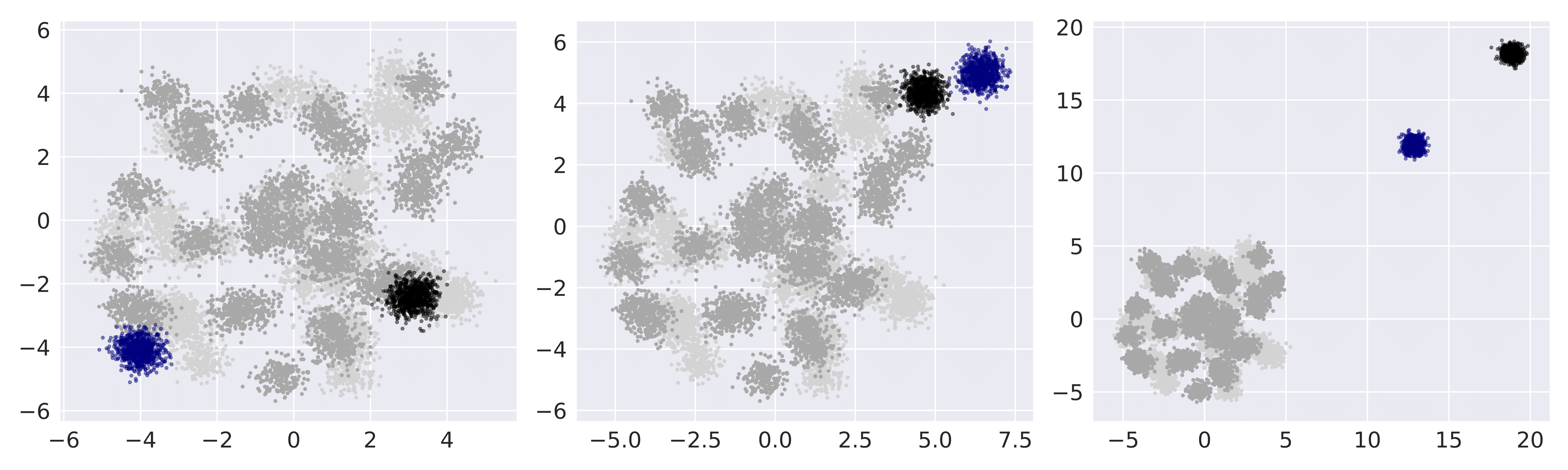}
\caption{Thirty mixtures drawn from the meta-distribution $\mathcal{M}$. We plot (in color) 3 unseen distributions whose parameters are drawn from (left) $U(-5, 5)$; (middle) $U(3,7)$; (right) $U(10, 20)$, the first two begin in and close to $\mathcal{M}$ whereas the last mixture is clearly outside of $\mathcal{M}$.}
\label{fig:mnist:gen}
\end{figure}

\begin{figure}[h]
\centering     
\subfigure[$\mu \sim U(-5, 5)$]
{\label{fig:c}\includegraphics[width=\linewidth]{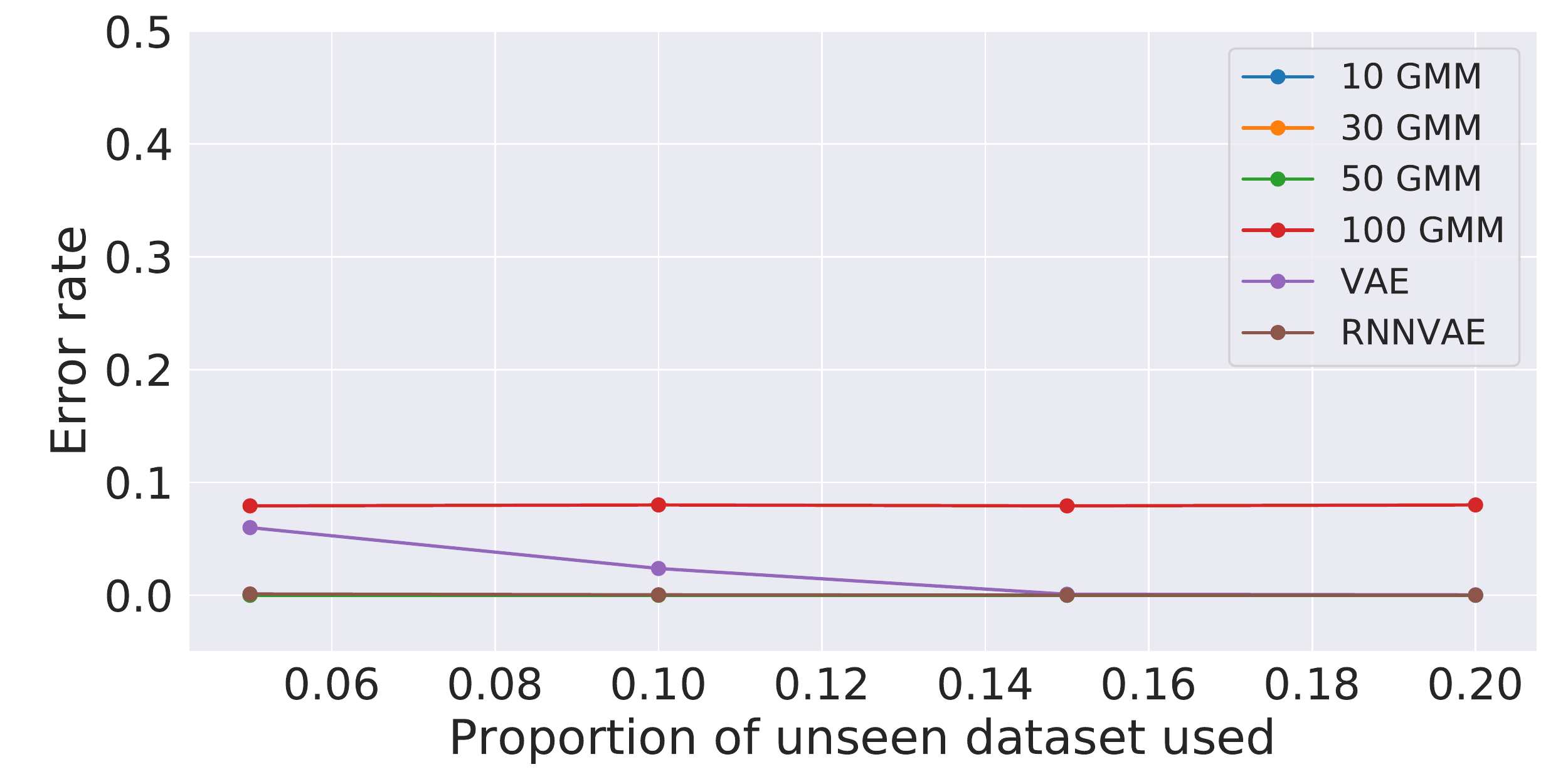}}
\subfigure[$\mu \sim U(3, 7)$]
{\label{fig:b}\includegraphics[width=0.49\linewidth]{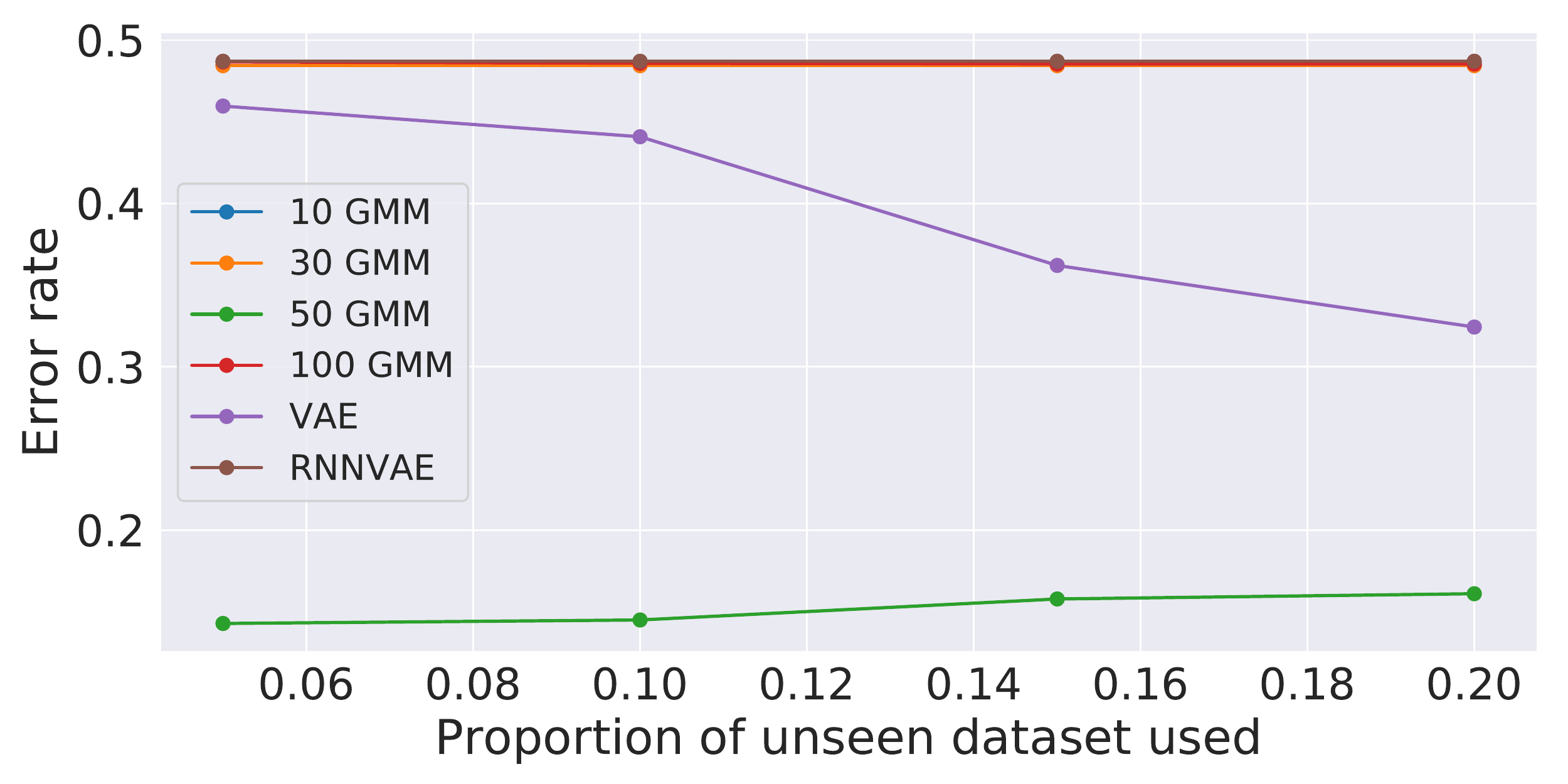}}
\subfigure[$\mu \sim U(10, 20$)] {\label{fig:a}\includegraphics[width=0.49\linewidth]{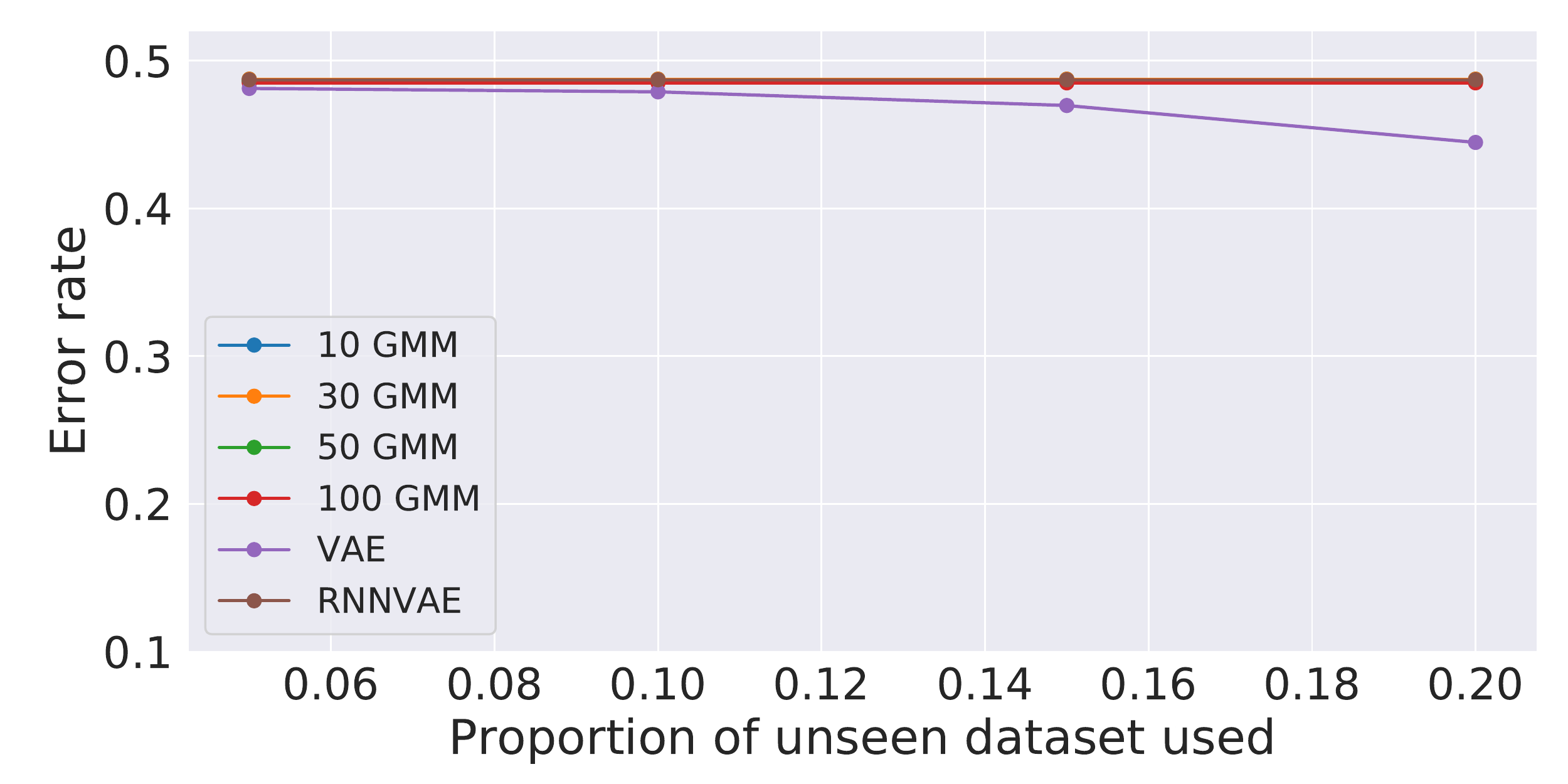}}
\caption{Clustering performance after training on \{5,10,15,20\}\% of the unseen data distribution. In (a), meta-training on 10, 30, and 50 datasets allows for perfect clustering, outperforming the VAE. In (b), only the 50 GMM meta-trained model has successfully learned to cluster. In (c), the meta-clustering algorithm fails to generalize to an extremely out-of-sample distribution.}
\label{fig:extra:mnist}
\end{figure}

Next, we investigate clustering ability of the meta-inference model on mixture distributions outside of $p_{\mathcal{M}}$ as we vary the amount of fine-tuning data (previously, we did not allow any fine-tuning -- inference was zero-shot). See Fig.~\ref{fig:mnist:gen} for different measures of generalizability. Specifically, we extract the pre-trained meta-inference model and train a new generative network on each of 3 unseen data distributions, evaluating the clustering performance. We only use \{5, 10, 15, 20\}\% of the test distribution for training. As shown in Fig.~\ref{fig:extra:mnist}(a), the model is able weakly generalize across all levels of meta-training, outperforming the VAE baseline with the exception of the 100 GMM meta-encoder -- a phenomena consistent with the results shown in Table 1, i.e., overfitting to the meta-training set. However, Fig.~\ref{fig:extra:mnist}(b,c) shows that meta-training does not seem to provide significant gains in generalization performance on marginals far from $p_{\mathcal{M}}$, again consistent with other demonstrations.

\section{Demo: Clustering Handwritten Digits}

Next, we construct a setup analogous to the mixtures of Gaussians experiment with MNIST digits \cite{lecun1998mnist}. Specifically, we hold out two digit classes for out-of-sample evaluation, and generate datasets comprised of pairs of the remaining digits. We select a subset of \{5, 10, 20\} combinations out of a total of 28 (8 choose 2) possibilities to train the MetaVAE. We then ask the model to cluster new digit pairs, either drawn from the eight unseen pairs in $\mathcal{M}$ or the digit pair (3s and 7s) that were held out completely in training. We use continuous 40-dimensional latent variables to better model the complexity of the data.

\begin{figure}[h!]
\centering
\subfigure[Digit Pair (1,6)]{\includegraphics[width=\linewidth]{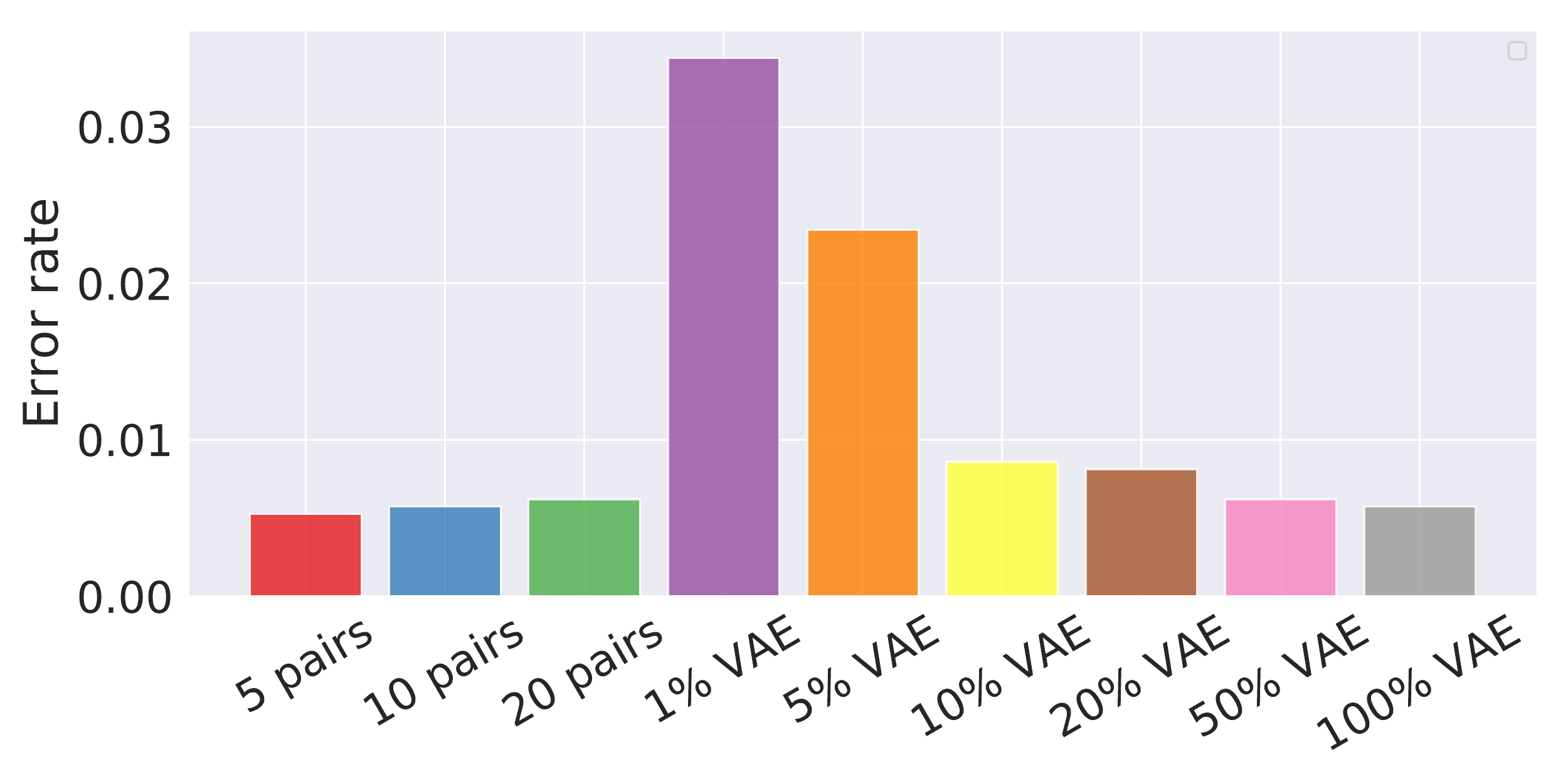}}
\subfigure[Digit Pair (4,9)]{\includegraphics[width=0.49\linewidth]{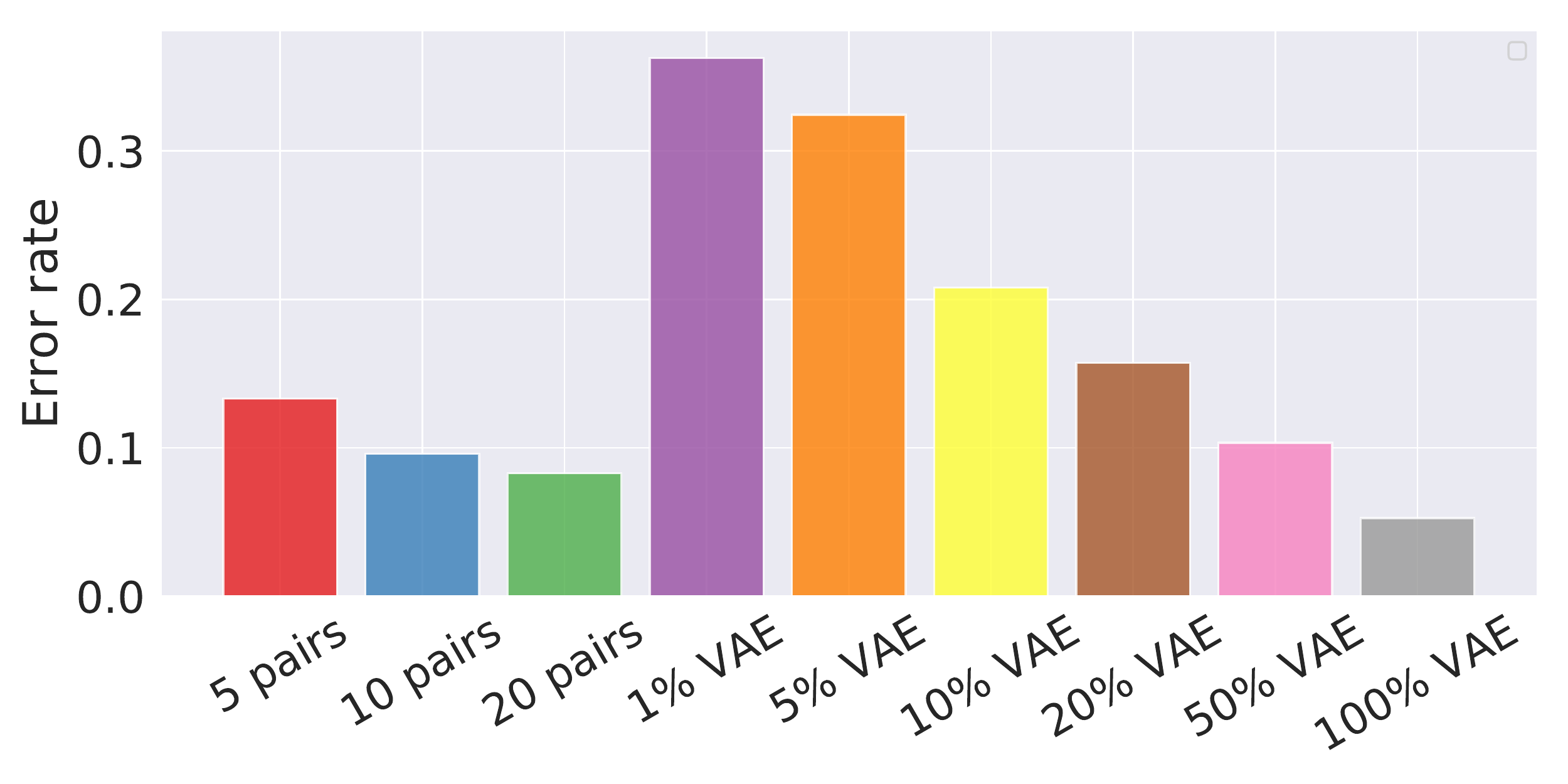}}
\subfigure[Digit Pair (3,7)]{\includegraphics[width=0.49\linewidth]{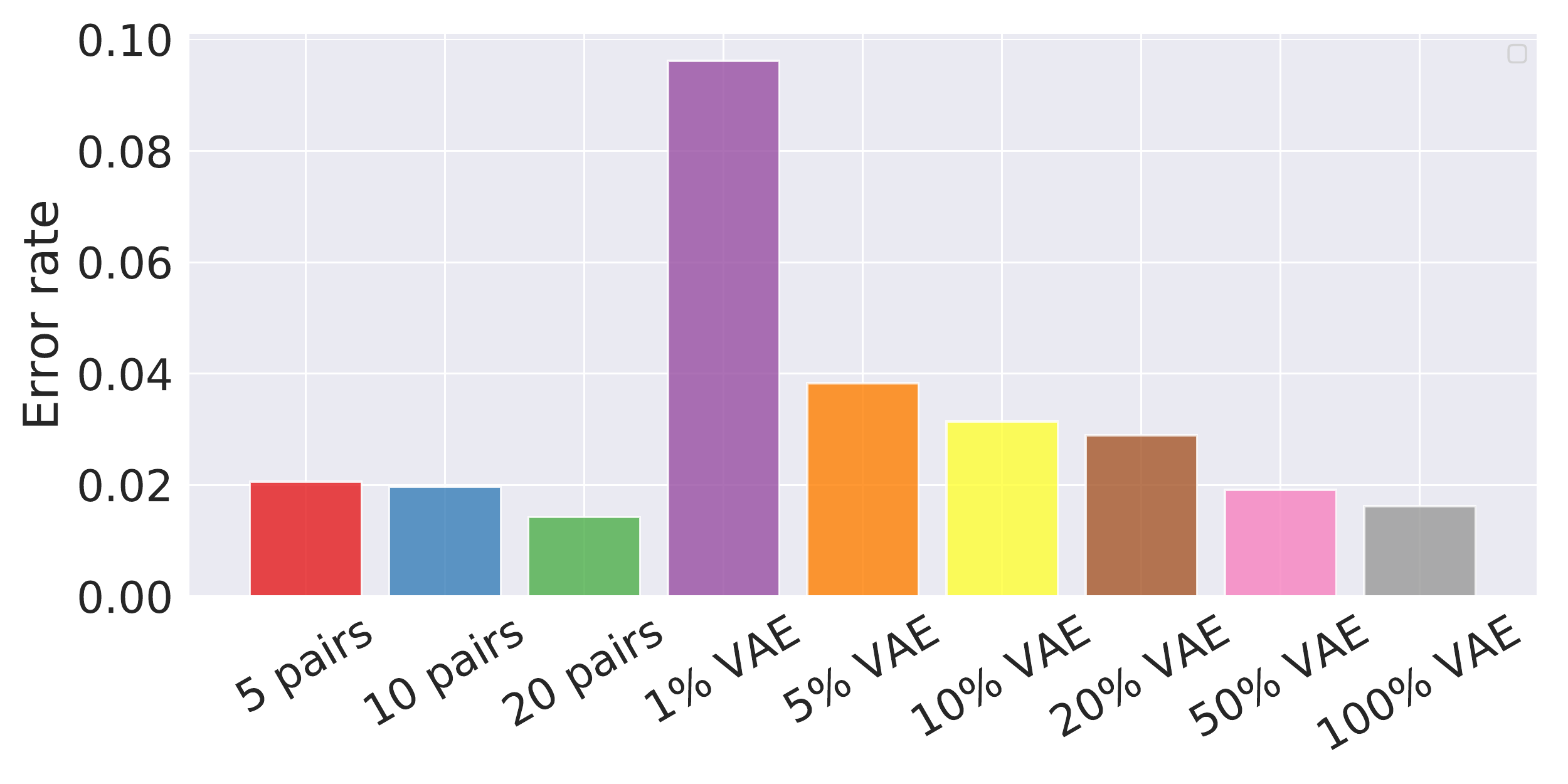}}
\caption{Clustering on MNIST digit pairs. We train a MetaVAE amortized over \{5, 10, 20\} pairs of digit classes and evaluate their performance on unseen pairs from and outside of  $p_\mathcal{M}$. (a,b) shows that the MetaVAE achieves higher clustering accuracy compared to a VAE trained on 100\% and 50\% of the target distribution (within $p_\mathcal{M}$). (c) shows that the MetaVAE outperforms a VAE trained on 100\% of the out-of-sample distribution (not in $p_\mathcal{M}$).}
\label{fig:mnist}
\end{figure}

Like in MoG, we use MetaVAE representations to train a logistic regression model with the true labels (0/1 for each digit class). 
To measure performance, we embed the test set and compare against true labels. Fig.~\ref{fig:mnist}(a,b) shows the clustering results for two levels of difficulty: digit pair (1,6) (visually easy) and (4,9) (visually hard). For the former, an MetaVAE outperforms the VAE trained on the \textit{full} dataset of 1's and 6's. For the more difficult task, adding more combinations improves clustering performance, and the MetaVAE outperforms a VAE trained on half of the target data. Fig.~\ref{fig:mnist}(c) shows MetaVAE performance on the out-of-sample digit pair (3,7). The MetaVAE obtains less than 2\% clustering error \textit{without additional gradient steps}. Further, surprisingly, it outperforms a VAE which has been trained on 100\% of the target dataset of 3's and 7's. 

\section{Demo: Classical Mechanics (Continued)}

We include the derivation for the meta-compiled inference objective from the main text. Note that this is very similar to \cite{le2016inference}.

\begin{align*}
    \mathcal{L}_\phi &= \mathbb{E}_{p_{\theta_i^*}(\vx)}[D_{\text{KL}}(p_{\theta_i^*}(\vz|\vx)) || g_\phi(\vz|p_{\theta_i^*}, \vx))] \\
    &= \int_{\vx} p_{\theta_i^*}(\vx) \int_{\vz} p_{\theta_i^*}(\vz|\vx)\log \frac{p_{\theta_i^*}(\vz|\vx)}{g_\phi(\vz|p_{\theta_i^*}, \vx)} d\vz d\vx \\
    &\propto \mathbb{E}_{p_{\theta_i^*}(\vx, \vz)}[-\log g_\phi(\vz|p_{\theta_i^*}, \vx)]
\end{align*}

\section{Demo: Distribution Statistics Details}
As this experiment setup is slightly involved, we provide a more thorough explanation with details here. 

Recall that a sufficient statistic is defined as a function $\phi(x)$ mapping realizations of a random variable to a vector in $\mathbb{R}^d$. We noted in the main text that for realizations of a ``random vector" (length $k$) whose entries each are a random variable distributed i.i.d. according to some exponential family, the sum $\sum_{i=1}^k \phi(x_i)$ of the sufficient statistics for realizations of each random variable in the vector. Finally, recall that the objective is: having seen many realizations of random vectors from different exponential family distributions, is it possible to learn a sufficient statistic for a new random vector that can be used to estimate the parameters of the (possibly unseen) underlying distribution that each random variable in the vector is distributed by?

If we treat an observation $\vx$ as a realization of a random vector, then the meta-inference model $g_\phi(p_{\mathcal{D}}, \vx)$, as a function of $\vx$, should act as a sufficient statistic for $p_{\mathcal{D}}$. A key distinction between the this experiment and the mixture of Gaussians (MoG) experiment is what an observation represents. In MoG, we represent the $i$-th observation $\vx_i$ as a 2-D vector sampled from a mixture distribution; when doubly amortizing, the meta-inference model $g_\phi$ takes as input $\vx_i$ and a marginal distribution, which we represent as a data set $\mathcal{D}_i = \{ \vx \}_i$. In contrast, in this experiment, the $i$-th observation is interpreted as a \textit{realization of a random vector} $\vx$. The meta-inference model $g_\phi$ still takes as input the observation and a marginal distribution. 

In this case, the marginal is a distribution over random vectors, which we represent as a set of realizations (samples) of random vectors. We studied four different cases (meta-distributions): \textbf{1) First,} we perform inference for all two dimensional Gaussian distributions with spherical covariance of 0.1 and a mean between -5 and 5. This implies that every random vector will be composed of i.i.d samples from a 2-D Gaussian distribution. The inference objective is estimate the unknown parameters of a new unseen Gaussian distribution after training. \textbf{2) Second,} we consider all two dimensional Log Normal distributions with spherical covariance of 0.1 and a mean between -5 and 5. \textbf{3) Third,} we consider all two dimensional Exponential distributions with scale less than 5. \textbf{4) Fourth,} we consider the union of distributions in the previous three cases (this defines the largest meta-distribution of the four cases). Note that each distribution defined above only has one free (continuous) parameter, which will serve as the statistic that we infer.

In each case, we must construct a meta-training and meta-test set where the former is used to train the MetaVAE and the latter is used to measure generalization of inference. To create the meta-training set, we randomly sampled 30 parameters defining 30 distributions (for example, sample 30 means from a uniform distribution $U(-5, 5)$ to define 30 Gaussian distributions). For each of the 30 distributions, we sample 20 times, building a 20-D random vector $\vx$. To represent the marginal distribution, we use a set of 10 random vectors, each sampled  i.i.d. For the meta-test set, we consider an interpolation of unseen distributions across a range of parameters. For example, for the first case of only amortizing over Gaussian distributions, we meta-test on Gaussians with means from -10 to 10 by 0.1 increments. By also considering means outside of -5 and 5 (the meta-distribution), we measure how well the MetaVAE can do inference in and outside of the meta-distribution. A similar design is used for cases 2 through 4. 

Next, we describe components of the MetaVAE. We place the full burden of learning onto the meta-inference model by making each generative model $p_{\theta_i}(\vx|\vz)$ parameter-free i.e. $g_\phi$ has no choice but to act as the sufficient statistic; Critically, this is possible since $p_{\theta_i}(\vx|\vz)$ is given the correct distributional family that $p_{\mathcal{D}_i}(\vx)$ belongs to (so it knows how to use $\vz$ to define a distribution). Knowing the correct distributional family also defines the loss function; for example, if we are given that $p_{\mathcal{D}_i}(\vx)$ is Gaussian, then $\vz$ represents the mean and we can use a Gaussian PDF in the lower bound computation. However, the meta-inference model $g_\phi(p_{\mathcal{D}_i}, \vz)$ is tasked with matching marginals with the correct families and must produce a latent variable $\vz$ to capture the parameters of the true distribution, $p_{\mathcal{D}_i}(\vx)$. Since the number (1) and dimensionality (2) of all sufficient statistics are identical, we can choose $\vz$ to be a two dimensional continuous random variable. Future work can explore more complex designs such as distributions with different numbers of sufficient statistics. For some statistics, we add a Softplus function to ensure that it is greater than 0 (e.g. scale for exponential distributions). In terms of architectures, we chose a multilayer perceptron (MLP) that ingests a set $\{\vx\}_i$ and outputs a set of hidden vectors that we average over into a single hidden vector. This network is used to reduce an observation (set of sample vectors from a distribution) into a single vector $\vh_i$ as well as the representation of the marginal distribution into a set of vectors, $\{\vh\}_i$. Together, $\vh_i$ and $\{\vh\}_i$ are ingested by a separate MLP to return variational parameters for the sufficient statistic. 

At test time, no additional training is needed to do inference for unseen distributions. For an unseen distribution, we use the meta-inference model as a  statistic to estimate the unknown parameter of the given distribution. We report the mean squared error against the true parameter of the underlying distribution, which is known when generating the dataset.

In each of the four cases, we compare our results to baseline models. When amortizing over a single family of distributions (e.g. cases 1 through 3), we compare an doubly-amortized inference procedure with a singly amortized one: we train a VAE on a distribution from the family with a randomly chosen statistic: $[-1.2, 1.1]$ mean for Gaussian, $[-0.5, 1.8]$ mean for Log Normal, $[1.4, 2.8]$ scale for Exponential. The goal of this baseline is to see how inference generalizes without amortizing over generative models (poorly as it turns out). For case 4, when considering multiple families from the Exponential families, we compare a MetaVAE amortized over 30 Gaussian, 30 Log Normal, and 30 Exponential distributions (for a total of 90 distributions) to three separate MetaVAEs, amortized over only 30 distributions of its family e.g. 30 Gaussians, 30 Log Normals, and 30 Exponentials respectively. Including these baselines again measures the effect of meta-amortization. Finally, in main text, we also tested how well inference works for other members of Exponential family that were not observed during training. To be specific, we included Weibull distributions with scale 1 and shapes from $[0,5]$, Laplace distributions with location 0 and scales in $[0,5]$, and ``symmetric" Beta distributions with two equal shape parameters from $[0,5]$.


\begin{figure}[h]
\centering     
\subfigure[]{\includegraphics[ width=0.49\linewidth]{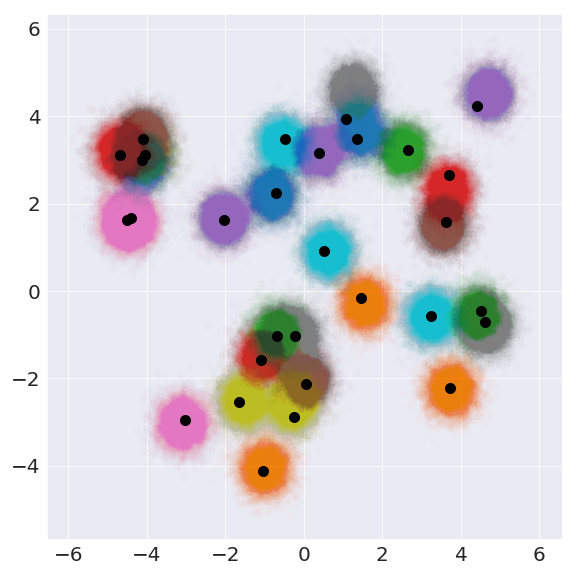}}
\subfigure[]
{\includegraphics[width=0.49\linewidth]{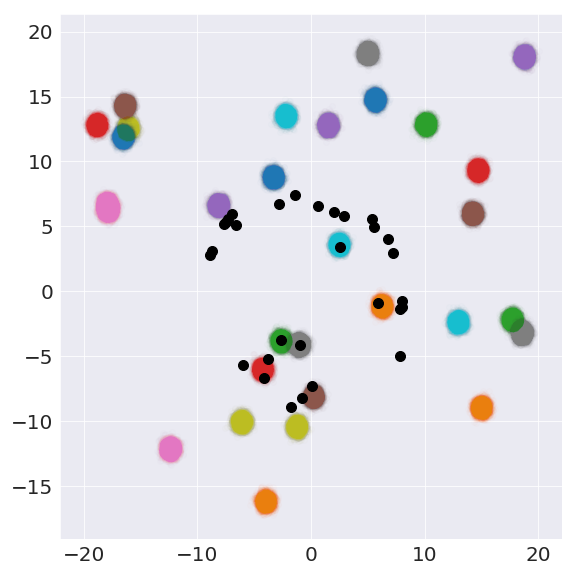}}
\subfigure[Log Normal]{\includegraphics[ width=0.49\linewidth]{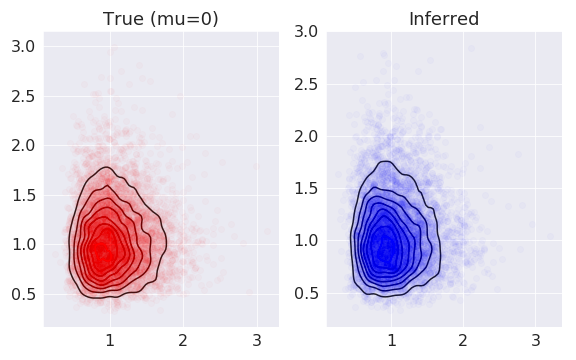}}
\subfigure[Exponential]
{\includegraphics[width=0.49\linewidth]{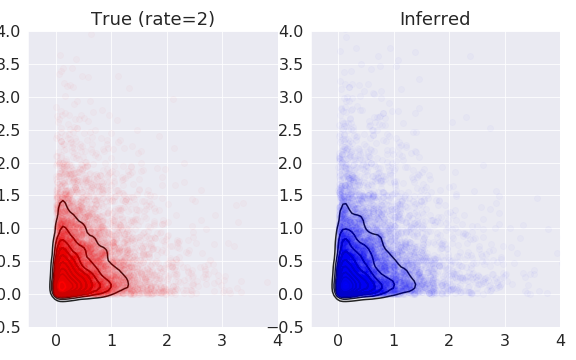}}
\caption{Colored circles represent 30 different $p_{\mathcal{D}_i} \sim p_{\mathcal{M}}$; black dots represent the inferred Gaussian means from the meta-inference model. (a) Test Gaussian distributions within $\mathcal{M}$; (b) Test distributions outside of $\mathcal{M}$. (c,d) Samples from the an unseen Log Normal or Exponential distribution $p_{\mathcal{D}_i} \in p_{\mathcal{M}}$ (red) and the true corresponding distribution defined by the inferred statistic (blue).}
\label{fig:gaussian_plot}
\end{figure}

\section{Training Details}

\subsection{Architectures}

In the main text, recall that $g_\phi(p_{\mathcal{D}_i}, \vx)$ is a supervised doubly amortized regressor that takes as input a marginal distribution $p_{\mathcal{D}_i}$ and an observation $\vx$ to return a posterior distribution. In practice, we need additional machinery to parameterize $g_\phi(p_{\mathcal{D}_i}, \vx)$ with neural networks. For some dataset $D_i$ and $\vx \in \mathcal{X}$, we set $\hat{g}_\phi(D_i, \vx) = r_{\psi}(\textsc{concat}(x, h_{\gamma}(D))$ where $\phi = \{ \psi, \gamma \}$, $h(\cdot)$ is \textit{summary} neural network that ingests the elements in $D$, and $r(\cdot)$ is an \textit{aggregation} neural network that ingests the input and the summary.

\paragraph{Mixture of Gaussians Experiment} The inference network for both the VAE and the MetaVAE is composed of 3 linear layers (hidden dimensions of 10) with ReLU nonlinearity in between each. The decoder networks share the same architecture as well. The summary network for the MetaVAE is also a MLP with three layers (hidden dimensions of 10) and Leaky ReLU nonlinearity.

\paragraph{Classical Mechanics Experiment} The inference model is identical to the MoG experiment except the latent variable is continuous (although still one-dimensional). No decoders are used as the simulators act as fixed generative models. The summary network is also as in MoG. 


\paragraph{Exponential Family Experiment} The inference network is composed 3 linear layers  (hidden dimensions of 400) with ReLU nonlinearity in between each. The summary network is also a MLP with three layers (hidden dimensions of 400) and Leaky ReLU nonlinearity. Results are not sensitive to choices of hidden dimension and nonlinearities.

\paragraph{MNIST and NORB Experiments} As many of the components as possible are shared between MetaVAE, NS, ad VHE. The latter two require additional sub-networks to ingest and decode a second (global) latent variables; thus, NS and VHE have more trainable parameters than MetaVAE. We use different designs for MNIST and NORB:

For MNIST, we use simpler architectures, flattening each image into a 784 dimensional vector. Specifically, we start with 3 linear layers with 400 hidden dimensions and ReLU nonlinearity for the encoder; 3 linear layers with 400 hidden dimensions and ReLU nonlinearity for each decoder; and 3 linear layers with 400 hidden dimensions and ReLU nonlinearity for the summary network. We used 40 latent dimensions (denoted $\vz$). For NS and VHE, we used an additional global latent (denoted $\vc$) of 300 dimensions and 3 linear layers with 400 hidden dimensions and ReLU nonlinearity to decode latent $\vz$ from latent $\vc$. 

Since NORB is more difficult (being realistic instead of synthetic images), we trade linear layers for convolutional architectures. Specifically, for the decoder, the MetaVAE uses: a linear layer first to increase the input dimensionality to $256*4*4$, which will be reshaped into an image; followed by six convolutional layers with three transposed convolutional layers every two convolutions with batch normalization after every layer (slowly decreasing the filter size from 256 to 128 to 64 to 1 or 3). For inference, the MetaVAE uses three sub-components: first, we have a large convolutional network with 9 convolutional layers with batch normalization in between layer that ingests the input image and outputs a object of size 256 by 4 by 4. Every input image and every sample from the distribution is processed using this convolutional network. Then the summary network consists of 3 linear layers with 400 hidden dimensions and ReLU nonlinearity that injests the output of the convolutional network into a summary statistic over samples. The resulting summary is concatenated with the output of the convolutional network for the input image and fed into two linear layers (400 hidden dimensions) with residual connections that spit out parameters of a Gaussian distribution over latent $z$. Again, VHE and NS have a second global latent variable of 300 dimensions that requires a separate decoder network, which we now define with two linear layers with residual connections (400 hidden dimensions).

\subsection{Hyperparameters}
\paragraph{Mixture of Gaussians Experiment} For the MetaVAE, we used a batch size of 20, a learning rate of 2e-4, and trained for 500 epochs using the Adam optimizer. For the VAE, we used a batch size of 100, a learning rate of 1e-3, and trained for 200 epochs using the Adam optimizer. The dataset was generated by sampling the appropriate MoG, where we sampled means uniformly from the ranges such as $U(-5, 5)$. We doubly-amortize over \{10, 30, 50\} such datasets at one time. We trained the model by exact enumeration of the ELBO/MetaELBO to avoid high-variance gradient estimates induced by using a 1-D discrete latent variable $\vz$.

\paragraph{Classical Mechanics Experiment} We use a batch size of 64, a learning rate of 2e-4, and trained for 10 epochs using Adam (for both VAE and MetaVAE). The dataset was created by running each simulator in the meta-train set 1000 times (similar for testing). For the VAE baseline we chose the ``center" simulator (a length of 10 in range 1 to 19 and an angle of 45 in range 5 to 85) which should give the best hope of generalization without doubly-amortizing. 


\paragraph{Exponential Family Experiment} We used a batch size of 20, a learning rate of 2e-4, and trained for 100 epochs using the Adam optimizer. The dataset was generated by sampling 1000 times i.i.d. from a parameterized distribution in the exponential family. We doubly-amortize over 10 to 30 such datasets at one time. The latent dimension was chosen to match the number of sufficient statistics and 20 i.i.d. samples where given to the summary network.

\paragraph{MNIST and NORB Experiments} We used a batch size of 100, a learning rate of 2e-4, and trained for 100 epochs using the Adam optimizer. MNIST images were kept at 28 by 28 pixels whereas NORB images were resized and center cropped to 32 by 32 pixels. All generative models used a latent dimension of 40 and 10 i.i.d. samples from the dataset to represent the distribution as input to $\hat{g}_{\phi}$.

\end{document}